\documentclass[10pt,twocolumn,letterpaper]{article}

\usepackage[pagenumbers]{cvpr} % To force page numbers, e.g. for an arXiv version

\usepackage{graphicx}
\usepackage{amsmath}
\usepackage{amssymb}
\usepackage{booktabs}
\usepackage{color}
\usepackage[usenames, dvipsnames]{xcolor}
\usepackage{arydshln}
\usepackage{pifont}% http://ctan.org/pkg/pifont
\newcommand{\cmark}{\ding{51}}%
\newcommand{\xmark}{\ding{55}}%

\definecolor{DeltaColor}{rgb}{0.039,0.73,0.71}
\definecolor{SetaColor}{rgb}{0.867, 0.0235, 0.376}
\definecolor{SigmaColor}{rgb}{0.98,0.45,0.0}
\definecolor{HaoColor}{rgb}{0.8,0,0}
\definecolor{AlphaColor}{rgb}{0,0,0.8}
\definecolor{BetaColor}{rgb}{0.8,0,0.8}
\definecolor{GammaColor}{rgb}{0.5,0,0.7}
\definecolor{EpsilonColor}{rgb}{0.353,0.725,0.906}
\definecolor{TauColor}{rgb}{0.0,0.0,0.8}

\usepackage[pagebackref,breaklinks,colorlinks]{hyperref}

\usepackage[capitalize]{cleveref}
\crefname{section}{Sec.}{Secs.}
\Crefname{section}{Section}{Sections}
\Crefname{table}{Table}{Tables}
\crefname{table}{Tab.}{Tabs.}

\def\cvprPaperID{4352} % *** Enter the CVPR Paper ID here
\def\confName{CVPR}
\def\confYear{2022}

\usepackage{overpic}
\usepackage{enumitem} %< control spacing in itemize/enumerate/...
\usepackage{overpic} %< add raw math symbols to figures
\usepackage{color}
\definecolor{turquoise}{cmyk}{0.65,0,0.1,0.3}
\definecolor{purple}{rgb}{0.65,0,0.65}
\definecolor{dark_green}{rgb}{0, 0.5, 0}
\definecolor{orange}{rgb}{0.8, 0.6, 0.2}
\definecolor{red}{rgb}{0.8, 0.2, 0.2}
\definecolor{darkred}{rgb}{0.6, 0.1, 0.05}
\definecolor{blueish}{rgb}{0.0, 0.3, .6}
\definecolor{light_gray}{rgb}{0.7, 0.7, .7}
\definecolor{pink}{rgb}{1, 0, 1}
\definecolor{greyblue}{rgb}{0.25, 0.25, 1}

\newcommand{\myparagraph}[1]{\noindent\textbf{#1}}
\newcommand{\boldparagraph}[1]{\vspace{0.05em}\noindent{\bf #1}.}
\DeclareMathOperator*{\argmin}{arg\,min}

\usepackage{blindtext}

\renewcommand{\paragraph}[1]{\vspace{1em}\noindent\textbf{#1}.}
\begin{document}
\title{COAP: Compositional Articulated Occupancy of People} % AB's suggestion for title. Please feel free to revert it back to original as you deem fit.
%\title{COAP: Compositional Articulated Occupancy of People for Robust Scene Interaction}
%\title{Articulation invariant neural implicit representation}

\author{
Marko Mihajlovic$^{1}$, Shunsuke Saito$^{2}$, Aayush Bansal$^{2}$, Michael Zollhoefer$^{2}$, Siyu Tang$^1$\\
$^1$ETH Z\"{u}rich \  $^2$Reality Labs Research \\[4pt]
  {\href[pdfnewwindow=true]{https://neuralbodies.github.io/COAP}{\nolinkurl{neuralbodies.github.io/COAP}}}\\[4pt]
}

\twocolumn[{%
    \renewcommand\twocolumn[1][]{#1}%
    \setlength{\tabcolsep}{0.0mm} %0

    \newcommand{\sz}{0.125}  % 0.125 0.11

    \maketitle
    \begin{center}
        \newcommand{\teaserwidth}{\textwidth}
    \vspace{-0.3in}
        \includegraphics[width=\linewidth]{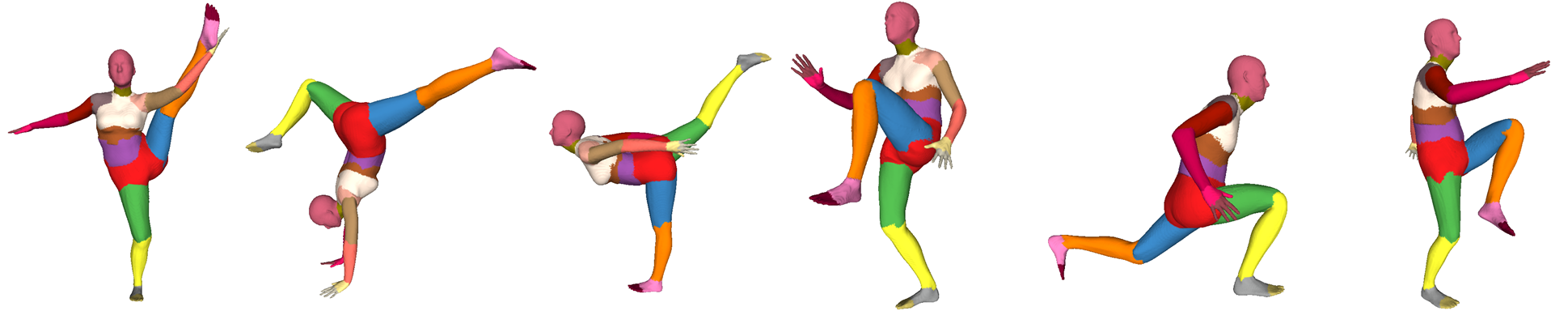}
      \vspace{-0.1in}
    %   \captionof{figure}{\textbf{Generalization on unseen shapes and poses.} We present \textbf{COAP}, a robust and generalizable neural occupancy representation for articulated human bodies. Compared to prior art such as LEAP~\cite{LEAP:CVPR:21} and Neural-GIF~\cite{NGIF}, COAP produces a much more reliable reconstruction on highly articulated out-of-distribution pose sequences from the PosePrior dataset~\cite{PosePrior_Akhter:CVPR:2015}.} 
        \captionof{figure}{We present \textbf{COAP}, a {\bf CO}mpositional {\bf A}rticulated occupancy representation of {\bf P}eople that is robust to highly articulated out-of-distribution pose sequences and that generalizes well to novel identities.
        The key idea is to decompose a human body into articulated body parts and employ a part-aware encoder-decoder architecture to learn neural articulated occupancy that models complex deformations locally. 
        The displayed example visualizes unseen subjects performing challenging poses from the PosePrior dataset~\cite{PosePrior_Akhter:CVPR:2015}. 
        The color scheme indicates labels of the activated articulated local implicit representations.
        % Compared to prior art such as LEAP~\cite{LEAP:CVPR:21} and Neural-GIF~\cite{NGIF}, COAP produces a much more reliable reconstruction on highly articulated out-of-distribution pose sequences from the PosePrior dataset~\cite{PosePrior_Akhter:CVPR:2015}.
        } 

    \label{fig:teaser}
    %\vspace{-0.2in}
    \end{center}%
}]
\begin{abstract}
We present a novel neural implicit representation for articulated human bodies.  Compared to explicit template meshes, neural implicit body representations provide an efficient mechanism for modeling interactions with the environment, which is essential for human motion reconstruction and synthesis in 3D scenes. 
However, existing neural implicit bodies suffer from either poor generalization on highly articulated poses or slow inference time. In this work, we observe that prior knowledge about the human body's shape and kinematic structure can be leveraged to improve generalization and efficiency. We decompose the full-body geometry into local body parts and employ a part-aware encoder-decoder architecture to learn neural articulated occupancy that models complex deformations locally. Our local shape encoder represents the body deformation of not only the corresponding body part but also the neighboring body parts. The decoder incorporates the geometric constraints of local body shape which significantly improves pose generalization. We demonstrate that our model is suitable for resolving self-intersections and collisions with 3D environments. Quantitative and qualitative experiments show that our method largely outperforms existing solutions in terms of both efficiency and accuracy.
% Neural implicit body modelling is a rising research topic in computer vision and graphics. 
% %
% Unlike traditional meshes, such 3D body representations enable efficient intersections with the environment which is essential for human motion reconstruction and synthesis in 3D scenes.
% %
% However, state-of-the-art neural bodies either suffer from poor generalisation on highly articulated poses or slow inference time. 
% %
% To address this problem, we propose a novel compositional neural implicit representation for articulated human bodies. 
% %
% We employ a simple encoder-decoder architecture.
% %
% The prior knowledge about human shape and kinematic structure are leveraged to design our local shape encoder that represents the shape deformation of not only the corresponding body part but also the neighbouring body parts.  
% %
% The decoder incorporates the geometric constraints of the local body shape which largely improves the pose generalisation. 
% %
% We demonstrate our model in 3 challenging down-streaming tasks, including resolving self-intersection, human body reconstruction in 3D scenes and body surface registration. Quantitative and qualitative experiments show that our method outperforms existing solutions both in terms of efficiency and accuracy.
\end{abstract}

\vspace{0.2cm}

\section{Introduction}
\label{sec:intro}

Computers can perceive rich representations of 3D human pose, shape, and motion by regressing the latent parameters of parametric human body models \cite{SMPL:2015, pavlakos2019expressive, xu2020ghum}. Conventionally, such generative human body models are represented as polygonal meshes and are easy to deform and animate by leveraging skinning algorithms such as linear blend skinning (LBS)~\cite{lbs}. However, they are not well suited for efficient interactions with 3D graphics environment and resolving self-intersections.
Unlike meshes, neural implicit representations~\cite{LEAP:CVPR:21, NGIF, chen2021snarf} are flexible, continuous, and support efficient intersection tests with the environment.
The state-of-the-art neural implicit body models~\cite{LEAP:CVPR:21, Saito:CVPR:2021, MetaAvatar:NeurIPS:2021} learn an inverse LBS network to convert an arbitrary point in 3D space to the canonical space where identity- and pose-dependent surface deformations are modeled.
While being effective in capturing surface deformations in the canonical space, the learned inverse LBS networks often suffer from poor generalization capability to highly articulated unseen poses (Fig.~\ref{fig:teaser}). 
% \ab{Is it possible to make a Figure-(2?) that shows the problem with these approaches? And also show how SNARF and the proposed method can help address it? And show the amount of time taken by SNARF to get the final output and time taken by proposed method? If so, we can connect it with the text}
%
SNARF~\cite{chen2021snarf} circumvents the need of learning the inverse LBS network by formulating the inverse mapping as a root-finding problem.
However, the model is learned per subject, and the computationally expensive root-finding prevents the practical application of their method for human body reconstruction in 3D scenes.
% 5) Proposed Approach (top-down description)
%To bridge the research gap, 
In this work, we present a novel part-aware encoder-decoder architecture that models compositional neural occupancy representations which are robust, efficient, and can generalize to a large variety of body shapes and highly articulated body poses. We name it COAP (\textbf{CO}mpositional \textbf{A}rticulated occupancy of \textbf{P}eople).

COAP is inspired by two key insights: First, the learned inverse LBS function in LEAP~\cite{LEAP:CVPR:21} captures spurious long-range correlations, making it hard to generalize to highly articulated unseen poses.
To address this, we get rid of the learned LBS and propose a novel local shape encoding that models the neural occupancy of articulated body parts by using a localized context of direct neighbors in the kinematic chain. 
% its own local shape information as well as its direct neighbors along the kinematic chain. 
This localized way of representing the body and its deformations reduces overfitting to the spurious correlations in the training set. 
%Breaking the global ties leads to better generalization to challenging poses at test time. 
Furthermore, given the local part encoding, the final whole human body is represented as a composition of these predicted local neural fields. Instead of a simple per-part combination as in NASA~\cite{nasa}, each local encoding in COAP contributes not only to the corresponding body part but also to the deformations of the neighboring body parts. 
Overall, the compositional neural fields modeled by the part-aware encoder-decoder architecture are effective and greatly benefit generalization (Sec.~\ref{sec:exp}). % as demonstrated in Sec.~\ref{sec:exp}. 
% We obtain artifact-free transitions between different body parts by learning to merge all available shape encodings for each part.

Second,  prior knowledge  about  human shapes that is carried by the parametric body models can significantly ease the task of learning robust neural representations. 
Similar to LEAP~\cite{LEAP:CVPR:21}, 
we use SMPL~\cite{SMPL:2015} as the starting point. 
Given the input bone transformations, we can effectively extract the relevant local body vertex positions. We leverage the per-part body vertices to create simple geometric primitives (such as 3D boxes) and incorporate them in the neural network architecture. This can be considered as a geometric prior of a local body shape which simplifies the learning problem and helps the neural network to properly allocate its modeling capacity around the surface. As demonstrated in our experiment, the effective fusion of the geometric prior and the learning power of neural networks is vital for the generalization capability of the learned representations.
%locally models complex deformations between joints.
%

% \ab{We systematically evaluate the robustness and representation power of our approach. COAP outperforms SNARF~\cite{}, which is trained per subject, both in accuracy and run-time. }
We systematically evaluate the robustness and the representation power of COAP. We compare with SNARF~\cite{chen2021snarf} that is trained per subject and shows impressive results on unseen poses~\cite{PosePrior_Akhter:CVPR:2015}. COAP achieves even better performance while at the same time being more efficient in terms of inference time. We also compare with LEAP~\cite{LEAP:CVPR:21} and Neural-GIF~\cite{NGIF} that produce generalizable neural implicit bodies. Once again, COAP significantly outperforms their results on the PosePrior~\cite{PosePrior_Akhter:CVPR:2015} and the DFaust~\cite{dfaust:CVPR:2017} datasets.

Resolving self-interpenetration of deformable 3D shapes is challenging and has been a long-standing question in computer graphics and vision \cite{guan2012virtual, pons2015metric, Bogo:ECCV:2016, pavlakos2019expressive, vaillant2013implicit, kavan2012elasticity, rohmer2009exact}. 
We propose a simple, yet effective optimization algorithm based on COAP that can efficiently resolve self-interpenetration among different body parts.
Our method can reliably solve the challenging cases that are not addressed by existing solutions \cite{pavlakos2019expressive} (as shown in Sec.~\ref{subsec:scenepen}). Furthermore, we demonstrate the utility of COAP for resolving collisions with 3D environments. 
Prior work~\cite{PROX:2019, zhang2020generating} requires pre-computed signed distance fields (SDFs) of 3D scenes to perform collision detection between 3D human bodies and the scene geometry, which is cumbersome and does not scale to scenes with moving objects or humans. 
Our robust and generalizable neural body model can be used to directly detect collisions with raw scans to improve 3D pose and shape estimation (Sec.~\ref{subsec:scenepen}). %by providing an accurate and efficient neural occupancy representation of 3D bodies and improves 3D pose and shape estimation results on the PROX dataset~\cite{PROX:2019}.

%We observe that prior knowledge about human shape and kinematic structure can provide better constraints for learning a representation that improves both generalization and efficiency. \figurename~\ref{fig:teaser} demonstrates the efficiency of our representation and its suitability for resolving self-intersections. 
%We use the SMPL~\cite{SMPL:2015} body model as the starting point and encode local shape information per body part by extracting the relevant vertices based on the skinning weights.
% 6) Impact/Advantages
%Given this novel local shape encoding, we predict the occupancy of each body part including its direct neighbors in separation.
%
%This localized way of representing the body and its deformation reduces overfitting to spurious correlations in the training set.
%
%Breaking these global ties leads to better generalization to challenging motions at test time.
%
%The final shape is represented as a composition of these predicted local neural fields.
%
%For artifact-free transitions between the different body parts, each local field models not only the corresponding body part alone but also neighboring deformations.
%
%This enables us to transition between the parts by smoothly blending all available predictions for each part.

% 7) What do we show in this paper
%In this work, we demonstrate the robustness of our method to novel motions and identities as well as its suitability to efficiently resolve challenging self-penetrations and collisions with other geometries. 

\paragraph{Contributions}
In summary, our main contributions are:
%
% 8) Contributions
% \boldparagraph{Contributions}s
%\begin{itemize}[itemsep=-4pt,topsep=2pt,leftmargin=*]
(1) a novel neural implicit body model that is robust and efficient, and can generalize to a large variety of human shapes and highly articulated body poses; 
(2) an effective localized encoder-decoder architecture that leverages local shape encoding and geometric shape priors to learn compositional neural body representations;
and, (3) simple and efficient optimization algorithms that reliably resolve challenging self-interpenetration and human-scene interpenetration. 
%We extensively analyze the representation power and evaluate the utility of the proposed neural body model. The surface reconstruction accuracy on unseen shapes and poses improves by 25\% on average comparing to the previous state-of-the-art methods. 
% 
Code and models are public\footnote{\href[pdfnewwindow=true]{https://neuralbodies.github.io/COAP}{\url{neuralbodies.github.io/COAP}}}. 
% \clearpage
\section{Related Work} \label{sec:related}
\subsection{Parametric Body Representations}
Parametric body models \cite{SMPL:2015, xu2020ghum, osman2020star, romero2017embodied} consist of a template mesh with an underlying kinematic skeleton. 
To animate a body, the canonical skeleton is reposed via forward kinematics and the mesh vertices are deformed by a skinning algorithm~\cite{kavan2005spherical, lewis2000pose, kavan2008geometric}. % with a set of artist-created skinning weights. 
Popular data-driven models such as SMPL~\cite{SMPL:2015} and GHUM~\cite{xu2020ghum} use the Linear Blend Skinning (LBS) algorithm to deform mesh vertices as a weighted sum of several rigid body part transformations.
% Such simple rigid transformations cannot model pose-dependent details which may result in a deformed mesh of a different volume. This problem is usually addressed manually by artists or automatically by adding pose-dependent correctives \cite{lewis2000pose, bailey2018fast}.
%
While human meshes are ubiquitous in computer graphics due to their good animation and rendering properties, they often self-intersect \cite{vaillant2013implicit} when a human body is reposed, and they are further not suitable for testing interactions with the environment. 
These two properties are essential for many human-scene interaction applications \cite{PROX:2019, LEMO:Zhang:ICCV:2021, rempe2021humor} and registration pipelines \cite{wang2021locally, bhatnagar2020ipnet} which often generate ill-defined models that self-intersect or collide with other objects. 
We address these two critical problems with our compositional neural implicit representation. 

\myparagraph{Resolving Self-intersections.}
Mesh self-intersection is a common problem in computer graphics that occurs when a human body mesh is reposed. 
To address this problem, most prior techniques~\cite{Li:2018:IOS, sifakis2007arbitrary, molino2003tetrahedral, nesme2009preserving} build an intermediate volumetric representation (\eg tetrahedral mesh) at every animation step and require an expensive optimization procedure to untangle self-intersecting bodies, which makes them unsuitable for image-based human reconstruction tasks~\cite{pavlakos2019expressive, ROMP, jiang2020coherent}. 
More efficient methods tailored for human bodies optimize human pose to resolve self-intersections. 
Guan~\etal~\cite{guan2009estimating, guan2012virtual} model each body part by their convex hull, which is in turn employed to create a differentiable penalty function for interpenetrated body parts. 
Since such an approach imposes a computationally expensive optimization problem, other works have proposed to alleviate the computation bottleneck by over-approximating body parts with simple geometric proxies (\eg spheres~\cite{pons2015metric} or capsules~\cite{Bogo:ECCV:2016}) to compute a differentiable interpenetration term efficiently. 
A more precise approach has been proposed in~\cite{Tzionas:IJCV:2016, pavlakos2019expressive}, which detects and penalizes self-intersected mesh triangles using a BVH tree~\cite{teschner2005collision}. 
However, such a loss term imposes a discretized surface-based error that is prone to local minima, whereas our method is volume-aware and imposes a more robust continuous penalty. 

\myparagraph{Resolving Collisions with the Environment.}
Modeling interactions of articulated parametric human bodies with raw scans or other geometries is a hard task. 
A common approach is to convert raw scans into meshes and penalize collided triangles \cite{Tzionas:IJCV:2016}. However, such methods impose a computationally expensive surface-based loss and are computationally expensive for more complex scenes. Hence most prior works \cite{PROX:2019, zhang2020generating, PLACE:3DV:2020, LEMO:Zhang:ICCV:2021} circumvent this problem by calculating SDF grids of raw scans, which is an error-prone task and not always possible~\cite{jacobson2013robust}.
Similarly, \cite{jiang2020coherent} propose to detect collisions between two human body meshes by dynamically calculating 3D SDF grids, which is memory and computationally expensive ($\approx$25s for $256^3$ grids) and erroneous when the meshes self-intersect. 
Our method circumvents these problems by representing a parametric human body as a volumetric representation that enables efficient differentiable collision checks with other geometries represented by meshes or point clouds. 
% As demonstrated in previous work \cite{LEAP:CVPR:21}, such implicit representations can straightforwardly model collisions with other humans that often appear in simulation \cite{zhang2020generating, PLACE:3DV:2020, hassan2021populating} and reconstruction pipelines \cite{jiang2020coherent, ROMP, fieraru2020three}. 

\subsection{Neural Implicit Representations.}
Neural implicit representations \cite{mescheder2019occupancy, park2019deepsdf, chibane2020implicit, peng2020convolutional, xie2021neuralfields} enable efficient inside/outside tests by representing shapes as signed-distance or occupancy functions parameterized by neural network weights. 
However, most of these representations are designed for rigid objects and cannot represent highly-articulated humans. 

\myparagraph{Neural Implicit Bodies.} 
Analogously to mesh-based body models, several recent works \cite{LEAP:CVPR:21, chen2021snarf, Saito:CVPR:2021, MetaAvatar:NeurIPS:2021, palafox2021npm, santesteban2021self} have proposed to learn neural implicit bodies. 
They simplify the learning problem by modeling neural representations in canonical space.  
NASA~\cite{nasa} learns a subject-specific part-based occupancy representation that is composed via rigid bone transformations in a posed space. 
However, the composition introduces artifacts around joints, and their low-dimensional pose encoding does not fully remove long-range spurious correlations. 
LEAP~\cite{LEAP:CVPR:21} and Neural-GIF~\cite{NGIF} propose to learn a generalizable neural implicit human body model in a canonical space and a separate inverse LBS neural network that projects any given query point to the canonical space where reliable occupancy checks are performed.
Similarly, SCANimate~\cite{Saito:CVPR:2021} and MetaAvatar~\cite{MetaAvatar:NeurIPS:2021} learn subject-specific avatars in a canonical space and an inverse LBS neural network to deform the surface points. 
These methods alleviate the problem of the artifacts around the joints presented in NASA\cite{nasa}. However, the learned inverse LBS is less robust to novel motions. 
imGHUM~\cite{alldieck2021imghum} employs a multi-part model and learns an implicit human representation directly in the posed space. 
% \siyu{rewrite the difference - we don't know whether it is hard or not. Also, they also use a part-based representation, we should clearly explain the difference}
%
SNARF~\cite{chen2021snarf} learns a subject-specific model in a canonical pose, but it circumvents the need for an inverse LBS network by formulating the inverse mapping as a root-finding problem. 
However, it suffers from computationally expensive inference and requires per subject training, which makes it less suitable for many practical applications. 
Compared to existing representations, our model better generalizes to novel motions and identities. %while being significantly faster for inference and training.
This is achieved by learning the implicit fields for articulated body parts and leveraging geometric priors and localized encoders that reduce the overfitting caused by spurious correlations. 

\begin{figure*}[t!]
\begin{center}
    \includegraphics[width=\textwidth]{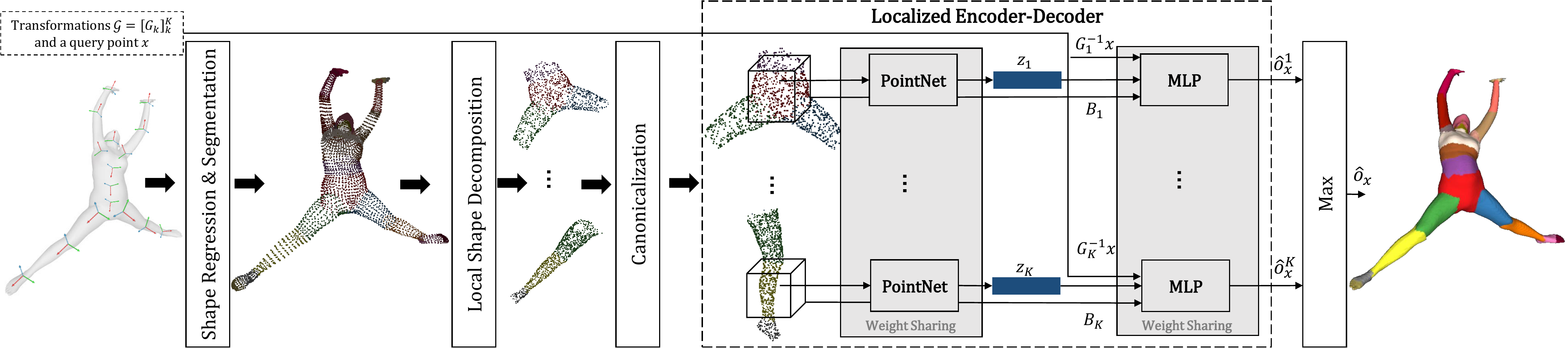}
\end{center}
\caption{
\textbf{Overview.}
We propose a part-aware neural network that consists of a localized shape encoder and decoder. 
The model takes the bone transformation matrices $\mathcal{G}$ as input and regresses (via SMPL~\cite{SMPL:2015}) a segmented human point cloud. 
The point cloud is then decomposed to articulated body parts which are canonicalized and encoded by the PointNet~\cite{qi2017pointnet} encoders. 
Finally, the decoder MLPs model each articulated body part independently as occupancy fields which are composed to represent the entire human body. 
The extracted mesh on the right represents the reconstructed human body, segmented according to the predictions of the decoder MLPs, for an unseen subject performing a novel pose from the PosePrior dataset ~\cite{PosePrior_Akhter:CVPR:2015}. 
% Our representation is a part-aware encode-decoder neural network that consists of a localized shape encoder and decoder. 
% The pipeline takes as input the bone transformation matrices $\mathcal{G}$ and regresses a global human shape represented as a sparse point cloud via SMPL~\cite{SMPL:2015}. 
% The global shape is then decomposed to articulated body parts and encoded by the PointNet~\cite{qi2017pointnet} encoders. 
% The decoder MLPs model each articulated body part independently as occupancy fields which are composed to represent a full human body. 
}
\label{fig:overview}
\end{figure*}

\section{Fundamentals} \label{sec:Fundamentals}
% We start by reviewing the fundamentals of parametric body models.

\boldparagraph{Modeling Human Bodies}
A parametric body model such as SMPL~\cite{SMPL:2015} is a data-driven model that is controlled 
via shape parameters $\beta$ and pose parameters $\theta \in \mathbb{R}^{K \times 3}$, where $K$ is the number of articulated joints.
It builds a human mesh in canonical pose $\bar{V}$ by deforming a pre-defined template mesh $\bar{T}$ via identity-dependent $\mathsf{B}_S(\beta)$ and pose-dependent $\mathsf{B}_P(\theta)$ vertex correctives:
\begin{equation} \label{eq:V_can}
    \Bar{V} = \Bar{T} + \mathsf{B}_S(\beta) + \mathsf{B}_P(\theta)\,.
\end{equation}
After this step, a skeleton composed of joint locations $\mathbf{J} \in \mathbb{R}^{K \times 3}$ in the canonical space is regressed by a learned matrix $\mathcal{J}$:
\begin{equation} \label{eq:joint_eq}
    \mathbf{J} = \mathcal{J}(\bar{T} + \mathsf{B}_S(\beta))\,.
\end{equation}
%
% \MZ{The skeleton is not dependent on the pose correctives?} \mm{Nope}

\boldparagraph{Reposing}
To animate a human body, the skeleton $\mathbf{J}$ in the canonical pose is reposed via the forward kinematics and can be compactly represented by a set of rigid bone transformation matrices $\mathcal{G}=[G_k]_{k=1}^K$ as 
\begin{equation} \label{eq:G_k}
G_k(\theta,\mathbf{J}) = \prod\nolimits_{j\in A(k)}
    \left[\begin{array}{@{}c|c@{}}
      R(\theta_j)
      & \mathbf{J}_j \\
    \hline
      \Vec{0} & 1
    \end{array}\right]\,,
\end{equation}
%
% \siyu{in the above equation, I would use $R_j(\theta)$}
where the rotation and the translation parts correspond to the bone orientation and the joint location, respectively. 
$R$ transforms pose parameters of the part $j$ into rotation matrix, and $A(k)$ defines a kinematic tree as an ordered set of ancestors of the joint $k$. 

Analogously to reposing of the canonical skeleton, the canonical mesh vertices $\Bar{V}$ are deformed via the linear blend skinning weights $\mathcal{W} \in \mathbb{R}^{K \times N}$ as a linear combination of rigid transformation matrices that define the mapping from the canonical to a posed space:
\begin{equation}\label{eq:v_i}
    V_i = \sum\nolimits_{k=1}^K \mathcal{W}_{k,i} G_k(\theta, \mathbf{J}) G_k(\Vec{0}, \mathbf{J})^{-1} \bar{V}_i\,,
\end{equation}
where $G_k(\Vec{0}, \mathbf{J})^{-1}$ removes the transformation due to the canonical pose (see \cite{SMPL:2015} for more details).

\boldparagraph{Shape Regression}
The bone transformation matrices $G_k$ define and fully constrain a human skeleton in a posed space. 
They encapsulate information about the canonical joints $\mathbf{J}$, which enables direct conversion of the transformation matrices into the shape vector $\beta$ for a small number of shape coefficients via the linear system:
\begin{equation}
    \mathsf{B}_S(\beta) = \mathbf{J} - \mathcal{J}\bar{T}\,.
\end{equation}

Such conversion enables us to interchangeably use the bone transformation matrices $G_k$ to represent shape coefficients $\beta$ and pose matrices $\theta$ and could be directly used to regress a human shape (Eq.~\eqref{eq:v_i}). 
In this work, we use the $G_k$-notation to be consistent with the previous neural body models~\cite{LEAP:CVPR:21, nasa}.

\section{COAP} \label{sec:representation}
COAP~(\textbf{CO}mpositional \textbf{A}rticulated occupancy of \textbf{P}eople) represents articulated human bodies as a differentiable implicit function. 
It defines the shape volume as the zero-level set $f_\Theta(x|\mathcal{G}) = 0$, in which $x \in \mathbb{R}^3$ is the input query point\footnote{Represented as homogeneous coordinates where appropriate.}, $\mathcal{G}=[G_k]_{k=1}^K  \in \mathbb{R}^{K \times 4 \times 4}$ is the input bone transformations with $K$ being the number of articulated body parts; 
% \siyu{add a footnote saying we have the same number of joints and body parts.}
we use the same number of articulated joints and body parts. 
% Our method is a neural model for representing articulated human bodies as a differentiable implicit function that defines shape volumetrically as zero-level set $f_\Theta(x|\mathcal{G}) = 0$, where $x \in \mathbb{R}^3$ is the input query point\footnote{for brevity, vectors represented as homogeneous coordinates where appropriate.}, $\mathcal{G}=[G_k]_{k=1}^K  \in \mathbb{R}^{K \times 4 \times 4}$ is the input bone coordinate frames, and $K$ is the number of articulated body parts. 

On a high level, our method first regresses the surface points of a human body using SMPL~\cite{SMPL:2015} and then implements a localized encoder-decoder neural network to represent human bodies as an implicit function. \figurename~\ref{fig:overview} shows an overview of our method. 
% The global shape is first decomposed to articulated body parts by leveraging human priors and then each part is processed independently via shared shape encoders and decoders. 
% The encoder module takes as input human bone transformation matrices $\mathcal{G}$ and converts them into latent codes  $[z_k]_{k=1}^K$ that represent $K$ articulated neural implicit body parts. 
% The decoder module takes these feature vectors as input and predicts whether a given query point $x \in \mathbb{R}^3$ is located inside the human geometry. 
% In the following, we describe these two modules. 

\subsection{Localized Shape Encoder}\label{sec:shape_encoder}
\paragraph{Body Shape Regression and Segmentation} 
The input bone transformations are first used to regress the deformed SMPL body vertices $V$ in the posed space (Eq.~\ref{eq:v_i}). %as described in Section~\ref{sec:Fundamentals}. 
These vertices $V$ are segmented to different body parts based on the SMPL skinning weights. 
%by associating each vertex with the label of the body part that has the largest skinning weight. 
% These vertices $V$ are then partitioned into $K$ articulated body parts, where each part captures complex shape deformations between joints. 

\paragraph{Local Shape Decomposition}
To encode the pose-dependent shape deformations, it is essential to consider not only the segmented body parts but also their neighborhoods in the kinematic chain.  
Therefore, for the local shape encoding of a segmented body part, besides its surface points, we also include the surface points that belong to its parent and child body parts in the kinematic chain. 
%We define the scope of articulated parts and their dependencies by using the skeleton ordering of the input transformation matrices and the introduced semantic labels in order to decompose the shape into articulated body parts. 
%Each articulated body part is defined to be dependent on several input bone transformation matrices. 
%Specifically, the part $k$ depends on the input bone $k$, its parent and its child bones
%all the predecessor bones 
%in the kinematic chain. 
Specifically, to compute the surface points for a segmented body part $k$, we use the skinning weights $W \in \mathbb{R}^{K \times N}$ and select all vertices in $V$ whose weights are larger than a threshold (empirically set to $0.01$) for all the body parts that are connected with the body part $k$.
We further extend this decomposition to mesh faces in the template mesh $\Bar{T}$, and sample points on the mesh surface. 
%Specifically, to compute surface points for the segmented body part $V^k$, we use skinning weights $W \in \mathbb{R}^{K \times N}$ and select all vertices in $V$ whose weights are bigger than a threshold (empirically set to $0.01$) for all the body parts connected in the articulated part $k$.
%We further extend this decomposition to mesh faces in the template mesh $\Bar{T}$. %which enables us to obtain local meshes for articulated body parts. 
% \paragraph{Local Dense Point Clouds}
Each local part is represented compactly with a point cloud as an intermediate representation. % by sampling points on the mesh surface of that part. 
More details about point sampling are in the supplementary materials. 
%
%Each point is sampled by first selecting a mesh face with probability proportional to the face area and then randomly sampling barycentric coordinates in order to calculate a point on the selected face. 

%To further balance the overlap among local articulated body parts, 
%the $k$th point cloud allocates one half of its capacity to encode the main component that corresponds to bone $G_k$, whereas the other half covers the whole local articulated body part region. 
%This design guides the neural networks to properly learn localized occupancy fields, where the largest part is reserved to represent the core bone component, while fewer samples for the non-central parts encourage smooth interpolation between connected occupancy fields. 

% projection to the canonical space 
\paragraph{Canonicalization}
Directly encoding the local point clouds as feature vectors makes learning hard since the neural networks need to reason about all possible human poses. 
Therefore, we simplify the learning problem by canonicalizing the point cloud of the local part  $k$ based on its bone transformation $G_k$. 
Let the $k$th point cloud be denoted as $\mathcal{P}^k$, then each point is projected to a canonical space via the corresponding bone transformation: 
\begin{equation}
    \hat{\mathcal{P}}^k_i = G_k^{-1} \mathcal{P}^k_i\,,
\end{equation}
where $\hat{\mathcal{P}}^k$ denotes the canonicalized point cloud of the body part $k$. 

\paragraph{Geometric Prior}
To further simplify the learning problem and help the neural networks properly allocate capacity, we build a simple geometric prior by constructing 3D bounding boxes $[B_k]_{k=1}^K \in \mathbb{R}^{K \times 6}$ for local body parts. 
These geometric primitives $B_k$ over-approximate the central component of the corresponding articulated body part and are estimated deterministically by finding extreme points in the local point clouds and adding an additional $15\%$ padding.  %\siyu{confusing, please rewrite} %and adding an additional padding of 15\%. 
%This geometric prior is later used in the occupancy decoder as an attention mechanism for the input query points. 

\paragraph{Local Shape Codes}
Canonicalized point clouds $\hat{\mathcal{P}}^k$ are then encoded via a PointNet~\cite{qi2017pointnet} as compact feature vectors $z_k \in \mathbb{R}^{128}$ that carry information about canonical shape and complex local deformations. 
These feature vectors are further augmented with one-hot encoding vectors for body parts to help the neural network learn a part-specific representation.
This localized PointNet encodes each articulated part independently and is implemented as a shared neural network for all articulated parts to reduce overfitting and improve the generalization to novel poses. 

\subsection{Neural Occupancy Decoder}
The second part of our approach %the auto-decoder pipeline 
is a decoder module that represents articulated body parts as occupancy fields which are composed to form a full human shape. 
The occupancy decoder takes as input the local shape codes $[z_k]_{k=1}^K$, the geometric prior $[B_k]_{k=1}^K$, the bone transformation matrices $[G_k]_{k=1}^K$, and a query point $x$ for which it predicts whether it is inside of a 3D human body. 

\paragraph{Local Occupancy Decoder}
First, the input query point $x \in \mathbb{R}^3$ is projected to the canonical space of the respective articulated body part $\hat{x}_k = G_k^{-1} x$. 
These local queries are augmented with a binary mask $b_k \in \{0, 1\}$ to facilitate the training by reducing the learning space, where $b_k$ indicates whether a local point $\hat{x}_k$ is inside of the created bounding box $B_k \in \mathbb{R}^6$. 
%Then, to facilitate the training by reducing the learning space, these local queries are augmented with a binary mask $b_k \in \{0, 1\}$ that indicates whether a local point $\hat{x}_k$ is inside of the created bounding box $B_k \in \mathbb{R}^6$. % This helps the neural models to allocate more capacity for points that are close to the surface. 
%
Next, the local query point $\hat{x}_k \in \mathbb{R}^3$, the binary mask $b_k \in \mathbb{R}$, and the local body code $z_k$ are concatenated as a feature vector and propagated through a 10-layer MLP that predicts occupancy value for the $k$th articulated part $\hat{o}_k$. 
The occupancy predictions are further multiplied by the weights $b_k$ to reduce potential spurious correlations. 
Similar to the local PointNet encoder, all local occupancy decoder MLPs share the same weights and perform occupancy checks independently to reduce overfitting. 
Please see the supplemental material for details about the neural network architecture. 

\paragraph{Occupancy Prediction}
The final occupancy prediction for the input query point is then determined as the union of localized occupancy predictions via the max operation:
\begin{equation}
    \hat{o}_x = \max [\hat{o}_k]_{k=1}^K\,.
\end{equation}
Note, there are two key differences between our approach and NASA~\cite{nasa} which also composes per-part occupancy representation to obtain occupancy prediction for full bodies.  
%Note that, although NASA~\cite{nasa} also composes per-part occupancy representations to obtain occupancy prediction for full bodies, there are several key differences compared to our model. 
First, our local shape encoding models a combination of local body parts and their direct neighboring parts along the kinematic chain, whereas NASA only captures single body parts. Second, we leverage shared occupancy decoders and geometric priors, while in NASA, each body part has an independent MLP, leading to poor generalization capability to out-of-distribution poses.

\subsection{Training}
We use the SMPL~\cite{SMPL:2015} body meshes from the AMASS dataset~\cite{AMASS:ICCV:2019} to train our model and the baselines.
For each body mesh in the training set, we sample a set of query points $P$. 
Half of these points are sampled uniformly inside the local bounding boxes $[B_k]_{k=1}^K$, while the other half is sampled around the mesh surface by using a Gaussian noise $x \sim \mathcal{N}(0, 0.1)$. 
For each query point, we compute the ground truth occupancy value $o_x \in \{0,1\}$ for supervision similar to the previous works~\cite{nasa, LEAP:CVPR:21} and activate the network output via the sigmoid function $\sigma$. 
Then, the final supervision loss is a simple mean squared error between the ground truth and the predicted occupancy values: 
\begin{equation} \label{eq:l2_training}
    \mathcal{L} = 
        \frac{1}{|P|} \sum\nolimits_{x \sim P} (\sigma(\hat{o}_x) - o_x)^2\,.
\end{equation}

We use the batch size of ten and optimize the model parameters via the Adam optimizer~\cite{kingma2014adam} with the learning rate of $10^{-4}$ and its default parameters. 
The representation fully converges after roughly 300k iterations for most experiments.

\begin{table*}
\centering
\scriptsize
\setlength{\tabcolsep}{3.4pt}

\begin{tabular}{@{}lcc|ccccc|ccccc@{}}
\toprule
\multicolumn{3}{l}{}        & \multicolumn{5}{c}{Female Subjects}  & \multicolumn{5}{c}{Male Subjects}      \\

Method                      & G      & t [ms] $\downarrow$  & 50004             & 50020             & 50021             & 50022             & 50025             & 50002             & 50007             & 50009             & 50026             & 50027             \\
\midrule
% NASA \cite{chen2021snarf}   & \xmark & xxx                  & 77.75/77.68       & 55.93/80.20       & 90.99/78.13       & 90.877/77862      & 71.20/78.64       & 68.14/74.82       & 67.57/71.82       & 44.84/74.32       & 87.44/77.47       & 48.84/79.30       \\
% Neural-GIF \cite{NGIF}      & \xmark & {\bf \phantom{0}22}  & xx.xx/xx.xx       & xx.xx/xx.xx       & xx.xx/xx.xx       & xx.xx/xx.xx       & xx.xx/xx.xx       & xx.xx/xx.xx       & xx.xx/xx.xx       & xx.xx/xx.xx       & xx.xx/xx.xx       & xx.xx/xx.xx       \\
% LEAP \cite{LEAP:CVPR:21}    & \xmark & \phantom{0}35        & 88.53/67.05       & 90.42/77.84       & 89.84/76.15       & 88.18/64.79       & 91.33/77.09       & 74.67/35.31       & 83.65/53.83       & 84.04/65.81       & 88.78/68.29       & 90.76/77.35       \\
SNARF \cite{chen2021snarf}  & \xmark & 809                  & 95.75/84.32       & 95.42/86.32       & 95.43/86.07       & 96.08/85.47       & 95.57/85.01       & 96.05/82.50       & {\bf 95.69/82.11} & 94.44/83.41       & 95.35/83.41       & 95.22/84.91       \\
COAP                        & \xmark & \phantom{0}{\bf 75}  & {\bf 95.97/85.35} & 95.84/87.62       & 95.57/86.82       & 95.98/85.65       & 95.84/86.28       & {\bf 96.61/82.96} & 95.27/81.90       & 94.91/84.90       & 96.07/85.89       & 95.78/86.90        \\%[2pt]
\midrule
% Neural-GIF~\cite{NGIF}      & \cmark & {\bf \phantom{0}22}  & 67.88/39.09       & 53.41/45.86       & 57.62/48.34       & 78.89/53.132      & 68.76/51.45       & 50.86/16.13       & 64.85/26.32       & 65.36/50.45       & 76.25/50.26       & 64.66/51.17       \\
% LEAP~\cite{LEAP:CVPR:21}    & \cmark & \phantom{0}35        & 88.53/67.05       & 90.43/77.85       & 89.84/76.15       & 88.18/64.79       & 91.33/77.09       & 74.66/35.30       & 83.65/53.82       & 84.04/65.81       & 88.78/68.29       & 90.75/77.35       \\
COAP                        & \cmark & \phantom{0}{\bf 75}  & 95.83/84.09       & {\bf 96.95/90.57} & {\bf 96.93/90.36} & {\bf 96.59/87.16}  & {\bf 97.24/90.36}& 86.75/58.75       & 93.89/76.72       & {\bf 96.16/88.15} & {\bf 96.79/88.22} & {\bf 96.89/89.97} \\
\bottomrule
\end{tabular}
\caption{
    \textbf{Single-subject neural implicit models.} Comparison of our model and SNARF~\cite{chen2021snarf} on subjects from the DFaust dataset \cite{dfaust:CVPR:2017} performing novel challenging poses from the PosePrior dataset \cite{PosePrior_Akhter:CVPR:2015}. 
    While both methods are quite robust, ours is over $10$ times faster and can additionally generalize to novel identities as shown in the third row;
    \textit{G} denotes whether the model has not seen test subjects during training; 
    values in cells are pairs of the mean IoU on uniformly sampled points and on points sampled around the ground truth meshes. 
}
\label{tab:sota_single_person}
\end{table*}

\begin{table}
    \centering
    \scriptsize
    \setlength{\tabcolsep}{8pt}
    \begin{tabular}{@{}lcc|cc@{}}
        \toprule
        \multicolumn{1}{l}{}                & \multicolumn{2}{c}{PosePrior Dataset \cite{PosePrior_Akhter:CVPR:2015}}  & \multicolumn{2}{c}{DFaust Dataset \cite{dfaust:CVPR:2017}}      \\
                                            & IoU Unif. & IoU Surf. & IoU Unif. & IoU Surf.      \\
        \midrule
        Neural-GIF \cite{NGIF}              & 65.83 & 58.21                         & 64.85 & 43.22      \\
        LEAP \cite{LEAP:CVPR:21}            & 89.36 & 73.33                         & 87.02 & 66.35      \\
        COAP                                & {\bf 96.97} & {\bf 89.92}             & {\bf 95.41} & {\bf 84.44}\\
        \bottomrule
    \end{tabular}
    \caption{
    \textbf{Generalization to unseen humans.} 
    Comparison of our model with LEAP~\cite{LEAP:CVPR:21} and Neural-GIF~\cite{NGIF} on the identities from the DFaust~\cite{dfaust:CVPR:2017} and the PosePrior~\cite{PosePrior_Akhter:CVPR:2015} datasets performing challenging novel poses from the PosePrior dataset; 
    values in the table correspond to the mean IoU of uniformly sampled points and of points sampled around the ground truth meshes respectively.
    }
    \label{tab:sota_unseen_ppl}
\end{table}

% \begin{table}
%     \centering
%     % \scriptsize
%     \setlength{\tabcolsep}{10pt}
%     \begin{tabular}{@{}lcc@{}}
%         \toprule
%         % \multicolumn{1}{l}{}              & \multicolumn{2}{c}{PosePrior Dataset~\cite{PosePrior_Akhter:CVPR:2015}}  & \multicolumn{2}{c}{DFaust Dataset~\cite{dfaust:CVPR:2017}}      \\
%                                             & PosePrior~\cite{PosePrior_Akhter:CVPR:2015} & DFaust~\cite{dfaust:CVPR:2017}     \\
%         \midrule
%         Neural-GIF \cite{NGIF}              & 65.83/58.21                         & 64.85/43.22      \\
%         LEAP \cite{LEAP:CVPR:21}            & 89.36/73.33                         & 87.02/66.35      \\
%         COAP                                & {\bf 96.97/89.92}                   & {\bf 95.41/84.44}\\
%         \bottomrule
%     \end{tabular}
%     \caption{
%     \textbf{Generalization to unseen humans.} 
%     Comparison of our model with LEAP~\cite{LEAP:CVPR:21} and Neural-GIF~\cite{NGIF} on the identities from the DFaust~\cite{dfaust:CVPR:2017} and the PosePrior~\cite{PosePrior_Akhter:CVPR:2015} datasets performing challenging novel poses from the PosePrior dataset; 
%     values in the table correspond to the mean IoU of uniformly sampled points and of points sampled around the ground truth meshes respectively. \siyu{can we indicate in the table that first number is for the uniform points and the second number is for close-to-surface points?}
%     }
%     \label{tab:sota_unseen_ppl}
% \end{table}

\section{Experiments}
\label{sec:exp}
We start by comparing our method with the state-of-the-art subject-specific neural implicit body model SNARF~\cite{chen2021snarf} and generalizable implicit body models, LEAP~\cite{LEAP:CVPR:21} and Neural-GIF~\cite{NGIF}, in Sec.~\ref{subsec:gen_rep_power}. Then, we conduct an ablation study to validate our design choices. 
% 
% in Sec.~\ref{subsec:gen_rep_power}. We compare our method with  SNARF~\cite{chen2021snarf}, a state-of-the-art neural implicit representation for subject-specific body models. We
% 
% We compare our method with the prior art in Sec.~\ref{subsec:gen_rep_power}. We compare our method with  SNARF~\cite{chen2021snarf}, a state-of-the-art neural implicit representation for subject-specific body models. We then compare with generalizable multi-person models, LEAP~\cite{LEAP:CVPR:21} and Neural-GIF~\cite{NGIF}. We then conduct an ablation study to validate our design choices. 
%
We further demonstrate the effectiveness of our representation to untangle self-intersected human bodies in Sec.~\ref{subsec:selfint} and study the benefit of COAP for estimating human-scene interactions \cite{PROX:2019} in Sec.~\ref{subsec:scenepen}. We conclude the section with a brief overview of the current limitations in Sec.~\ref{subsec:limitations}.
% \sh{commented out limitation section as it's currently missing.}

\subsection{Generalizable Representation Power} \label{subsec:gen_rep_power}

\paragraph{Experimental Setup}
For a fair comparison with the baselines, we assume a human skeleton topology with 24 body parts ($K=24$ in Sec.~\ref{sec:representation}) and use the DFaust \cite{dfaust:CVPR:2017}, MoVi~\cite{ghorbani2020movi} and PosePrior\cite{PosePrior_Akhter:CVPR:2015} datasets to train and evaluate our representation. 
% 
%As an evaluation metric, 
We report the \textbf{mean inference time} in ms for 10k points, the mean Intersection Over Union (\textbf{IoU}) of uniformly samples query points in a bounding box around the ground truth mesh, and the IoU of points sampled around the ground truth surface ($\mathcal{N}(0, 0.01)$) \cite{LEAP:CVPR:21, chen2021snarf}.%, similar to the previous work \cite{LEAP:CVPR:21, chen2021snarf}. 

\paragraph{Single-subject Neural Implicit Models} 
We start by comparing our method with SNARF~\cite{chen2021snarf}, a state-of-the-art subject-specific neural implicit body representation. 
Both methods are trained for each subject in the DFaust dataset \cite{dfaust:CVPR:2017} and evaluated on the challenging poses from the PosePrior dataset \cite{PosePrior_Akhter:CVPR:2015}. 
%Results shown in \tablename~\ref{tab:sota_single_person} demonstrate 
We observe in \tablename~\ref{tab:sota_single_person} that both methods are robust for challenging poses, whereas ours is more than \textbf{10 times faster} while being more accurate in most scenarios. 
Our method additionally generalizes to novel identities and motions. As demonstrated in \tablename~\ref{tab:sota_single_person} (3rd row), our model that is trained on MoVi~\cite{ghorbani2020movi} sequences can be directly used for DFaust subjects with challenging poses and produces even higher accuracy than the per-subject trained models from SNARF~\cite{chen2021snarf} in \tablename~\ref{tab:sota_single_person} (1st row).

\paragraph{Generalization to Unseen Subjects}
We now compare our model with two recently proposed neural body representations, LEAP~\cite{LEAP:CVPR:21} and Neural-GIF~\cite{NGIF}, which generalize to unseen identities. We train our model on the MoVi~\cite{ghorbani2020movi} dataset and use the pretrained baselines provided by the authors; LEAP trained on the same MoVi dataset, and Neural-GIF on augmented multi-shape SMPL models. As validation datasets, we use novel identities from the PosePrior~\cite{PosePrior_Akhter:CVPR:2015} and the DFaust~\cite{dfaust:CVPR:2017} datasets and sample novel poses from the challenging PosePrior dataset. 

Quantitative results are displayed in \tablename~\ref{tab:sota_unseen_ppl} (see Sup. Mat. for qualitative results) and demonstrate that our method significantly outperforms the baselines in terms of accuracy. 
This robustness comes from the compositional design of our representation and not requiring an inverse LBS network that poorly generalizes to novel motions. 
% In addition to robustness, the inference time of our representation is three times faster because our model is implemented as localized MLPs without additional deformation modules which are required by the baselines. 
This further enables faster and end-to-end training, whereas the baselines employ multi-stage training for the LBS networks that is less stable and more sensitive to hyperparameter tuning. 

In summary, our implicit representation is efficient, fast, and robust for articulated human bodies. 

\paragraph{Ablations Study}
\begin{table}
\centering
\scriptsize
\setlength{\tabcolsep}{2.5pt}
\begin{tabular}{@{}cc|cc@{}}
\toprule
Geometric Prior  & One Hot Encoding  & IoU Local Boxes  [\%] $\uparrow$  & IoU Surface  [\%] $\uparrow$  \\
\midrule
                &                   & 91.99             & 82.81         \\
                & \cmark        & 92.14             & 84.46         \\
\cmark      &                   & 92.99             & 85.44         \\
\cmark      & \cmark        & \textbf{93.61}    & \textbf{86.86}\\

\bottomrule
\end{tabular}
\caption{\textbf{Ablation study} quantifies the impact of the geometric prior $b_k$ and the one hot encoding vectors for the local features (Sec.~\ref{sec:representation}); 
\text{IoU Local Boxes} and \text{IoU Surface} are respectively the mean IoU of points uniformly sampled around bounding boxes $B_k$ and points sampled around the ground truth surface. 
The metrics are computed on the PosePrior sequences~\cite{PosePrior_Akhter:CVPR:2015}. 
}
\label{tab:ablation}
\end{table}

%Lastly, we provide a short ablation study to quantify the i
Lastly, we study the impact of the geometric prior and the one-hot encoding vectors (Sec.~\ref{sec:representation}) in Tab.~\ref{tab:ablation}. 
All methods are trained for 200k iterations on the MoVi dataset and evaluated on the PosePrior sequences. 
We observe that using both geometric prior and one-hot encoding improves the accuracy of our model.

% \begin{figure}
%     \begin{center}
%         \includegraphics[width=\linewidth]{figures/selfpenloss.pdf}
%     \end{center}
%     \caption{\textbf{Surface-based [] vs our volume-based self-intersection loss term.} TODO: sketch figure}
%     \label{fig:selfpen}
% \end{figure}

\subsection{Resolving Self-intersections} \label{subsec:selfint}
%Unlike previous neural implicit body models~\cite{LEAP:CVPR:21, NGIF, chen2021snarf} that model humans as holistic implicit fields that cannot be straightforwardly used to resolve self-intersection, our compositional body model naturally offers this ability and is robust for challenging cases. 
Prior work on neural implicit bodies~\cite{LEAP:CVPR:21, NGIF, chen2021snarf} models humans as a holistic implicit field. 
Such modeling restricts them from straightforwardly resolving self-intersections. On the contrary, our compositional body model naturally offers this ability and is robust for challenging cases. 

\paragraph{Method}
% Here we present an effective optimization procedure that can be used to resolve such problems. 
Given the self-intersected human body parameters as input (\eg SMPL shape $\beta$ and pose $\theta$ vectors, see Sec.~\ref{sec:Fundamentals}), we seek the optimal human pose $\theta^*$ such that the human body does not self-intersect. 
% The input parameters are first encoded as the shape codes $[z_k]_{k=1}^K$ as described in Sec.~\ref{sec:shape_encoder}. 
%
We take inspiration from the traditional computer graphics methods \cite{ericson2004real, thiery2013sphere} that use geometry proxies to efficiently approximate collisions. We propose to use 3D boxes to approximate body parts in order to efficiently detect potential collided body parts. 
Based on these collided boxes, we compute their intersected volumes in which we uniformly sample an initial set of points. 
From this initial set, we select only a subset of points that are inside of at least two body parts by checking our part-wise occupancy predictions. 
Let this final set be denoted as $\mathcal{S}$, then our self-intersection loss term is defined as:
\begin{equation} \label{eq:self_pen_loss}
    \argmin_{\theta} \sum\nolimits_{x \in \mathcal{S}} \sigma\left(f_\Theta(x|\mathcal{G})\right)\,.
\end{equation}

To further prevent unnecessary pose distortions (common in prior approaches \cite{PROX:2019, LEMO:Zhang:ICCV:2021}), we explicitly disable detecting collisions between kinematically connected body parts that almost always intersect.
Please see the supplementary material for additional implementation details. 

% First, our implicit representation takes as input (human shape $\beta$ and pose $\theta$) 
% Given the input SMPL parameters (human shape $\beta$ and pose $\theta$), our representation encodes them as a latent human encoding $\mathcal{Z}$  as described in Section~\ref{}. 

% \textit{Second}, inspired by traditional computer graphics methods \cite{ericson2004real, thiery2013sphere} that use geometry proxies to efficiently approximate collisions, we use a set of boxes to over-approximate body parts. 

% \textit{Third}, we detect pairs of collided boxes and calculate intersection volume
% \textit{Fourth}, we randomly sample a set of query points inside of the intersected volumes. 
% Then, by using our part representation, we select a subset of points $\mathcal{S}$ that penetrate at least two body parts by checking per part occupancy and propose an efficient self-intersection loss term (Eq.~\ref{eq:self_pen_loss}) that penalizes self-intersections.
% 
% \begin{equation} \label{eq:self_pen_loss}
%     \argmin_{\theta} \sum_{x \in \mathcal{S}} \sigma\left(f_\Theta(x|\mathcal{G})\right)\,.
% \end{equation}
%
% To further prevent unnecessary pose distortions, we explicitly disable collisions between directly connected body parts to prevent unnecessary distortions as common in the related methods \cite{PROX:2019, LEMO:Zhang:ICCV:2021}.

\begin{figure}
    \begin{center}
        \includegraphics[width=1.0\linewidth]{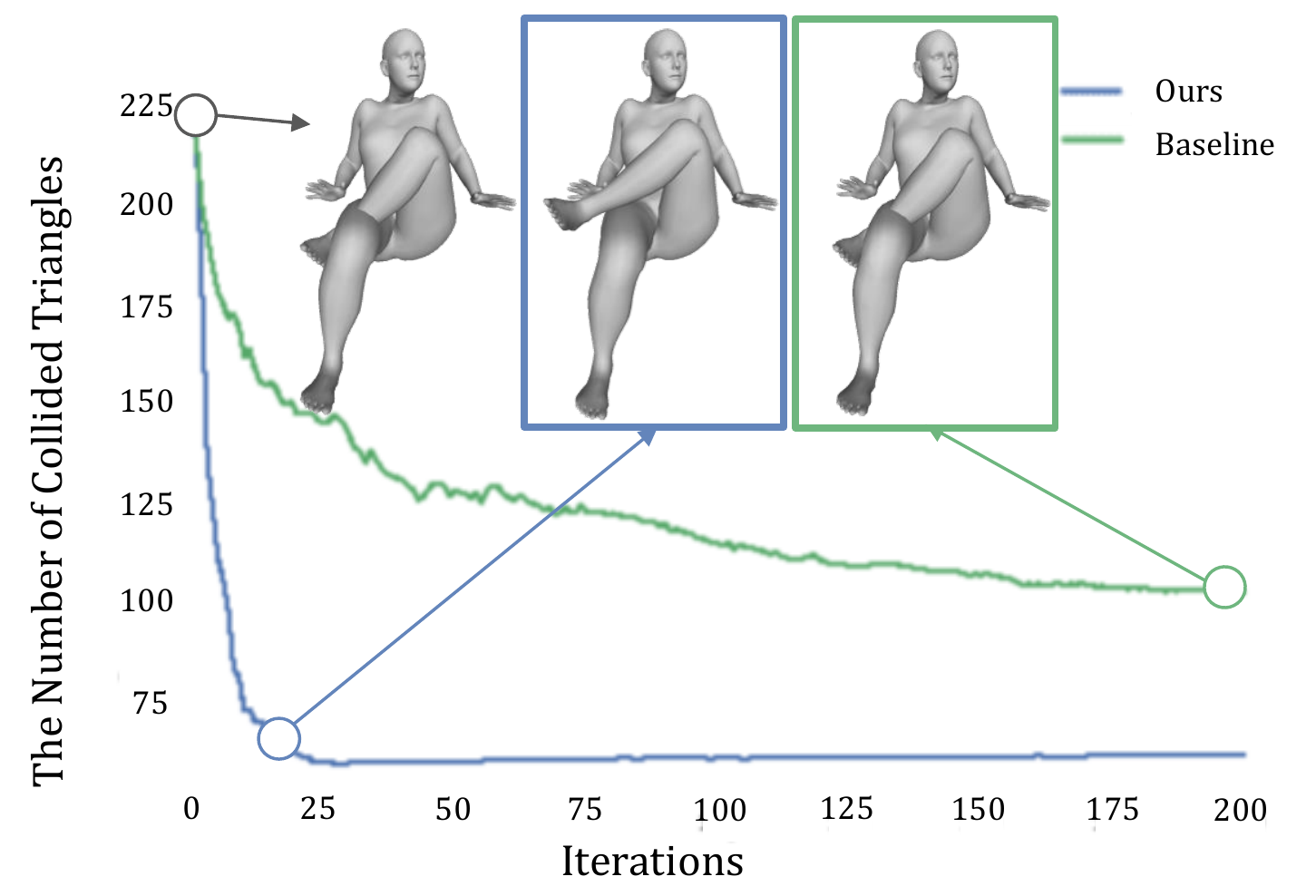}
    \end{center}
    \caption{\textbf{Resolving self-intersected humans} from the PROX~\cite{PROX:2019} dataset. 
    Our method successfully resolves the challenging self-intersections and converges faster compared to the baseline by a large margin \cite{Tzionas:IJCV:2016, pavlakos2019expressive}.}
    \label{fig:self_pen_chart}
    \end{figure}
    
\begin{figure*}[h!]
    \scriptsize
    \setlength{\tabcolsep}{0.0mm} %0
    \newcommand{\sz}{0.110}  % 0.33
    \begin{tabular}{ccc|ccc|ccc} % for solid lines
    % \begin{tabular}{ccc:ccc:ccc} % for dashed lines
        Initial pose   & \cite{pavlakos2019expressive, Tzionas:IJCV:2016}   & Ours & Initial pose   & \cite{pavlakos2019expressive, Tzionas:IJCV:2016}   & Ours & Initial pose   & \cite{pavlakos2019expressive, Tzionas:IJCV:2016}   & Ours \\
        \includegraphics[width=\sz\linewidth]{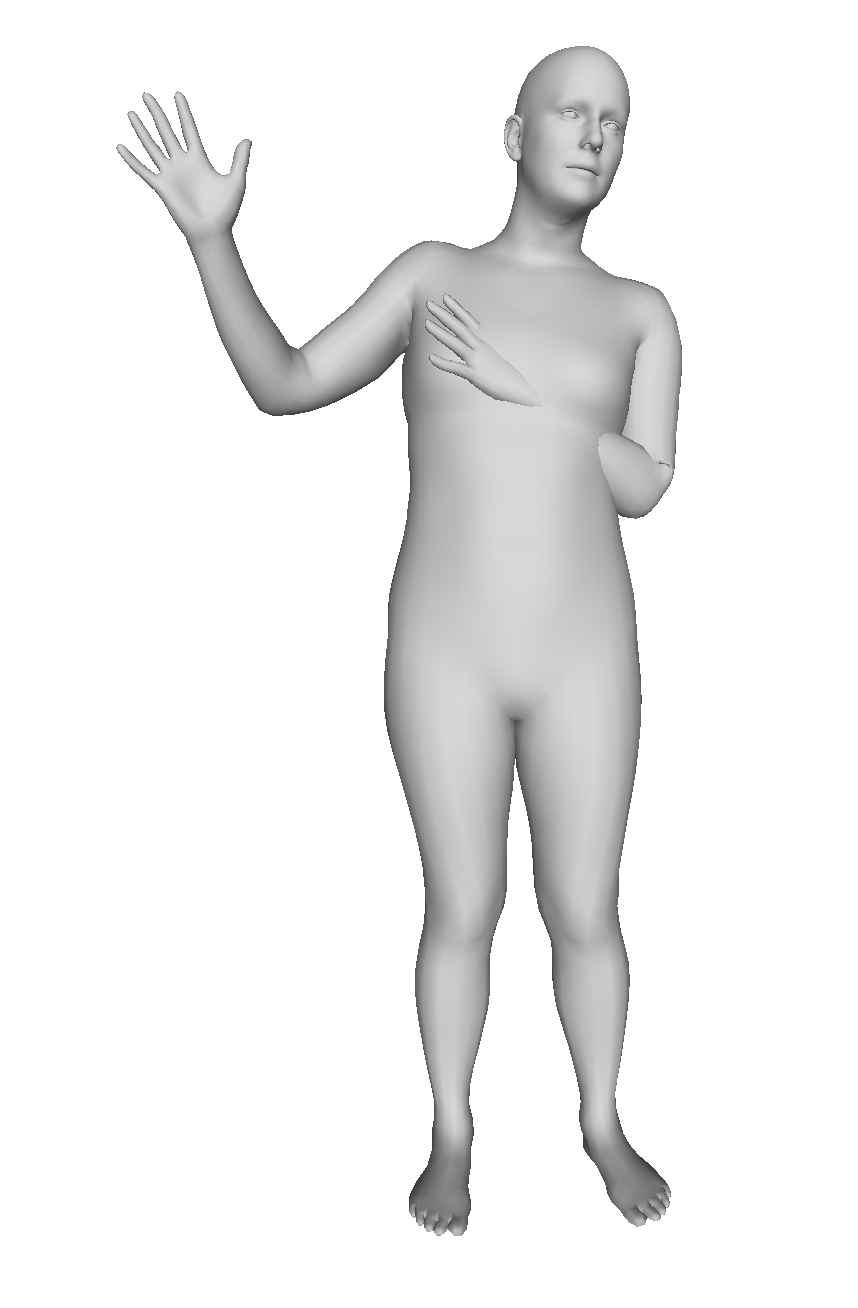} & \includegraphics[width=\sz\linewidth]{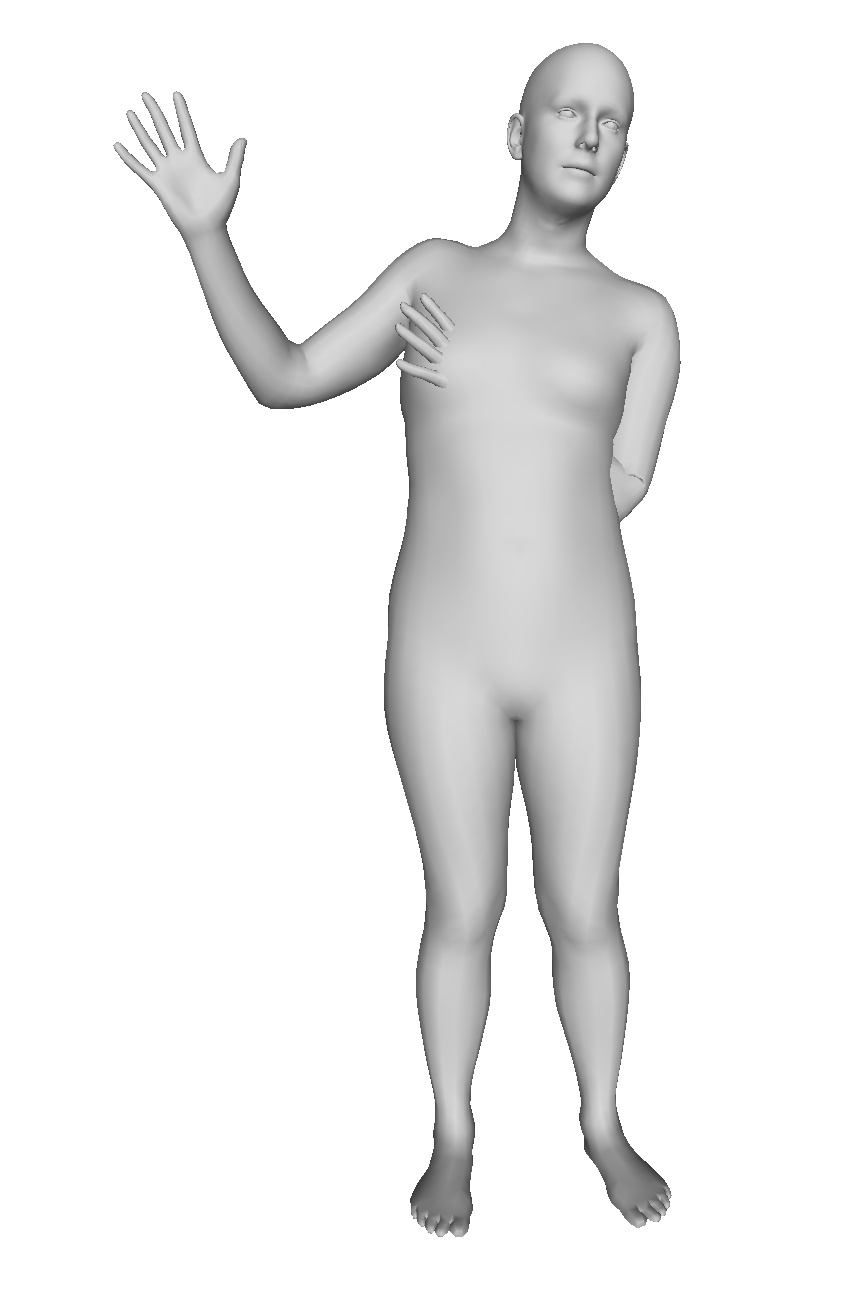} & \includegraphics[width=\sz\linewidth]{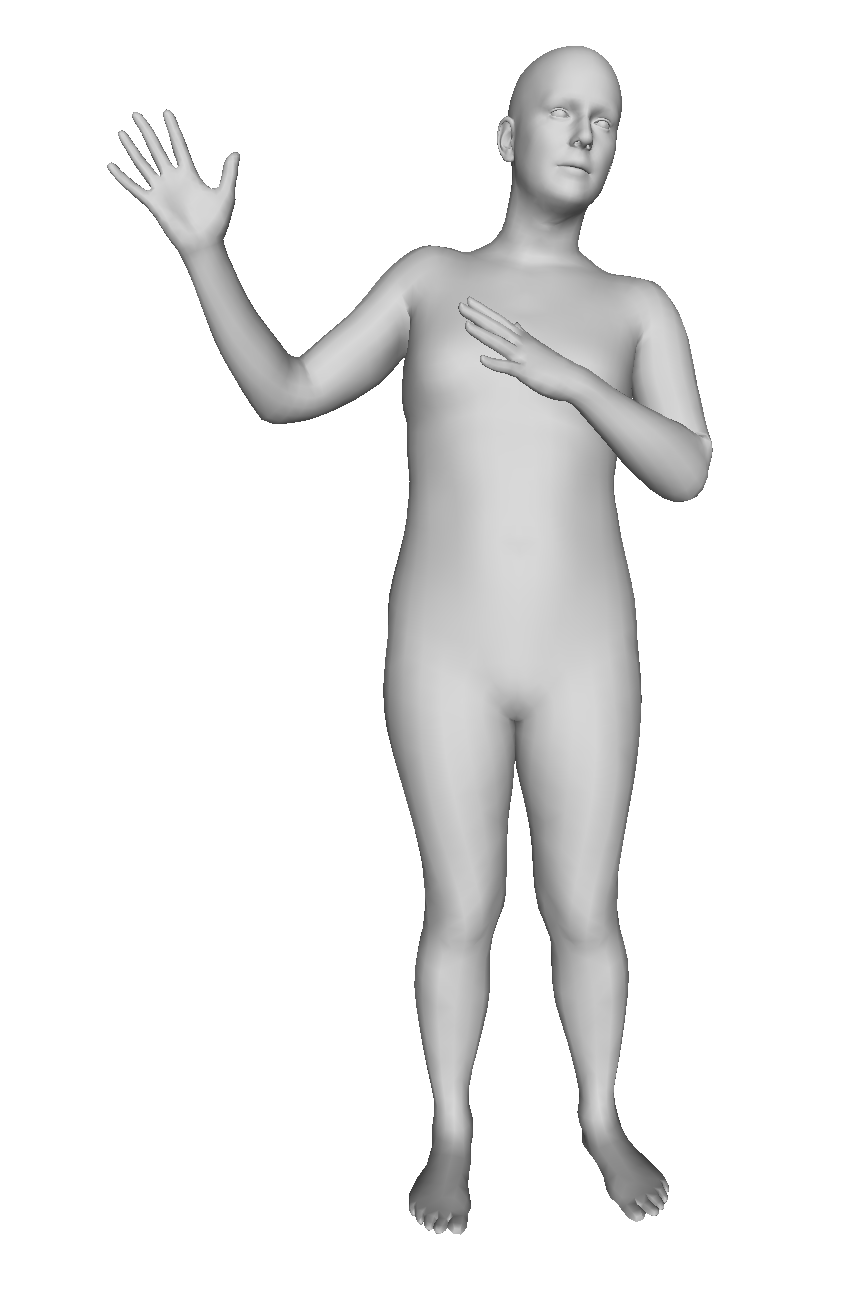} &
        
        \includegraphics[width=\sz\linewidth]{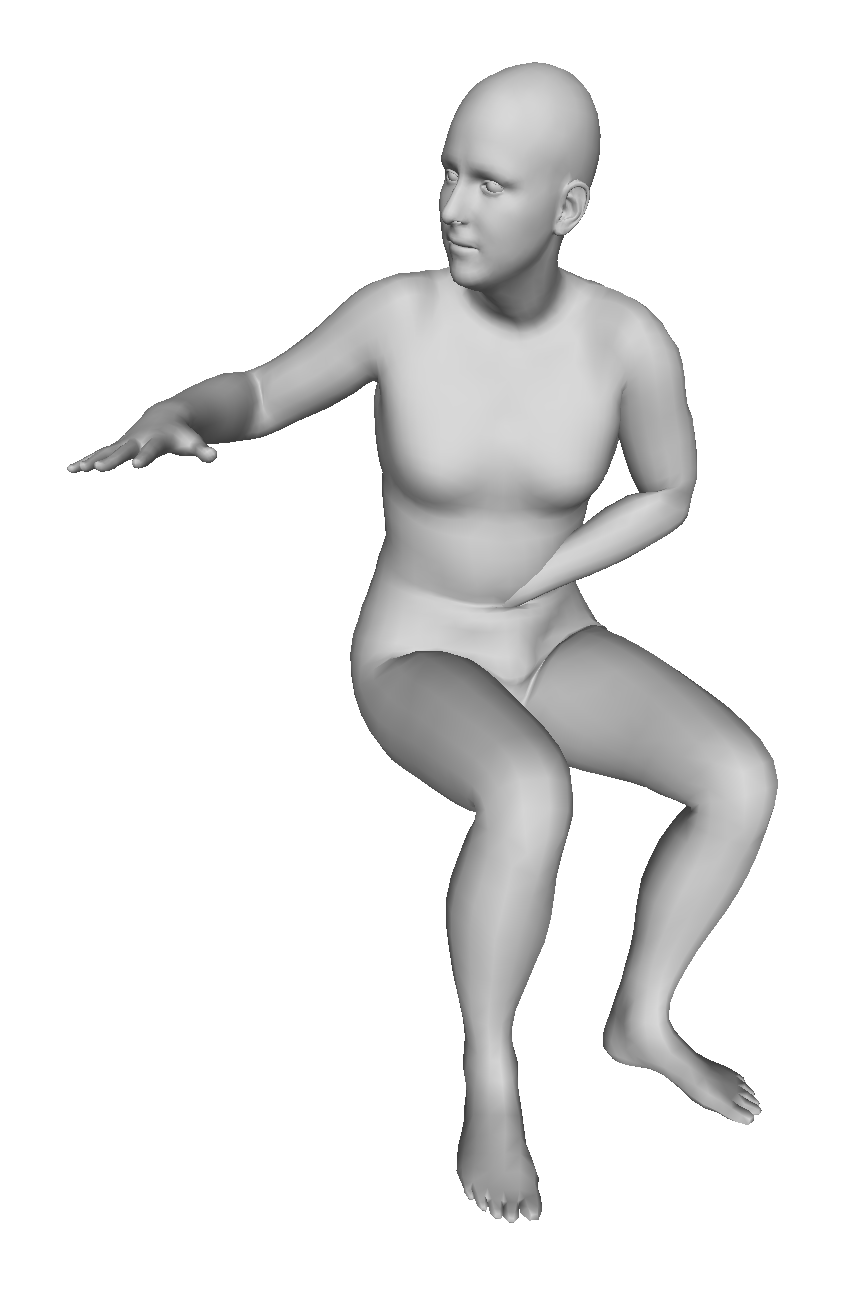} &
        \includegraphics[width=\sz\linewidth]{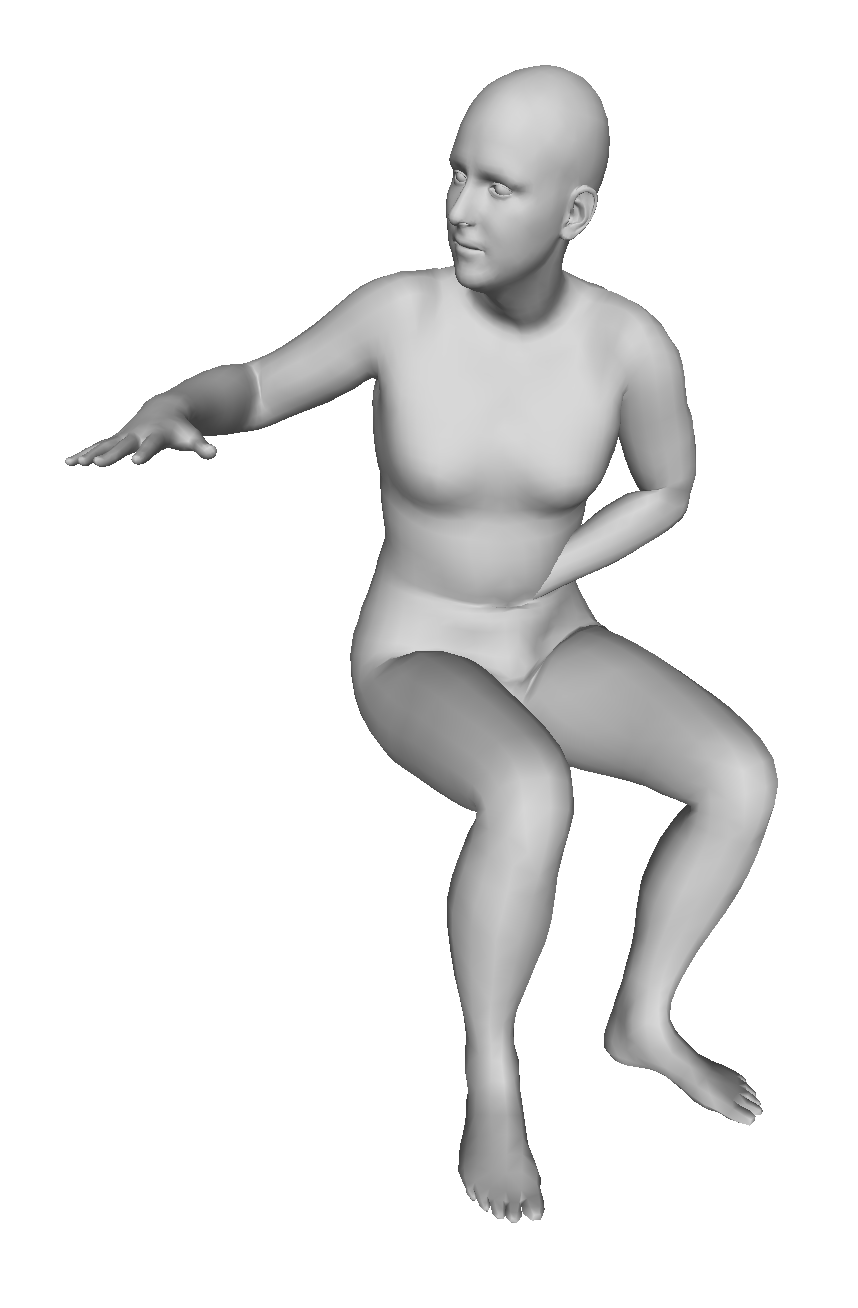} &
        \includegraphics[width=\sz\linewidth]{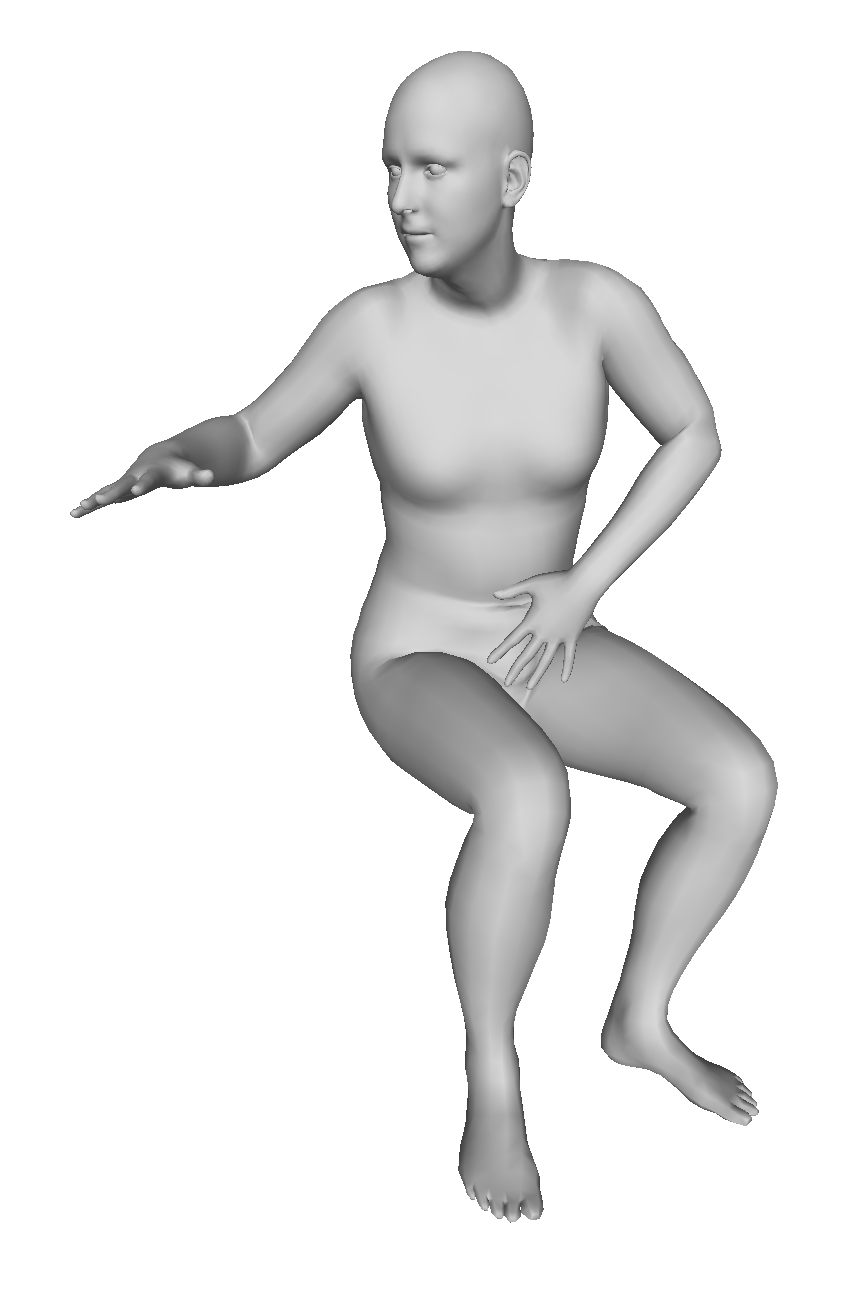} &

        \includegraphics[width=\sz\linewidth]{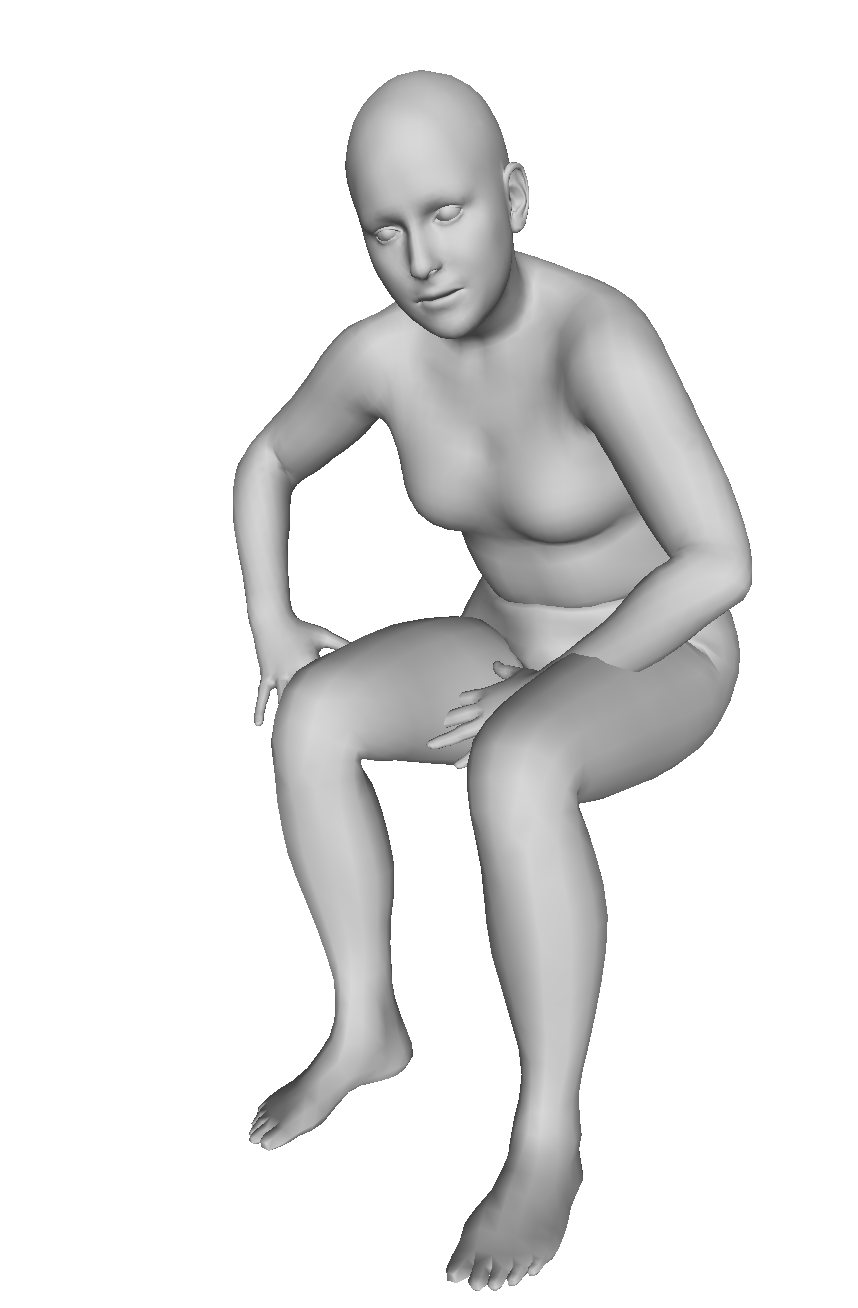} &
        \includegraphics[width=\sz\linewidth]{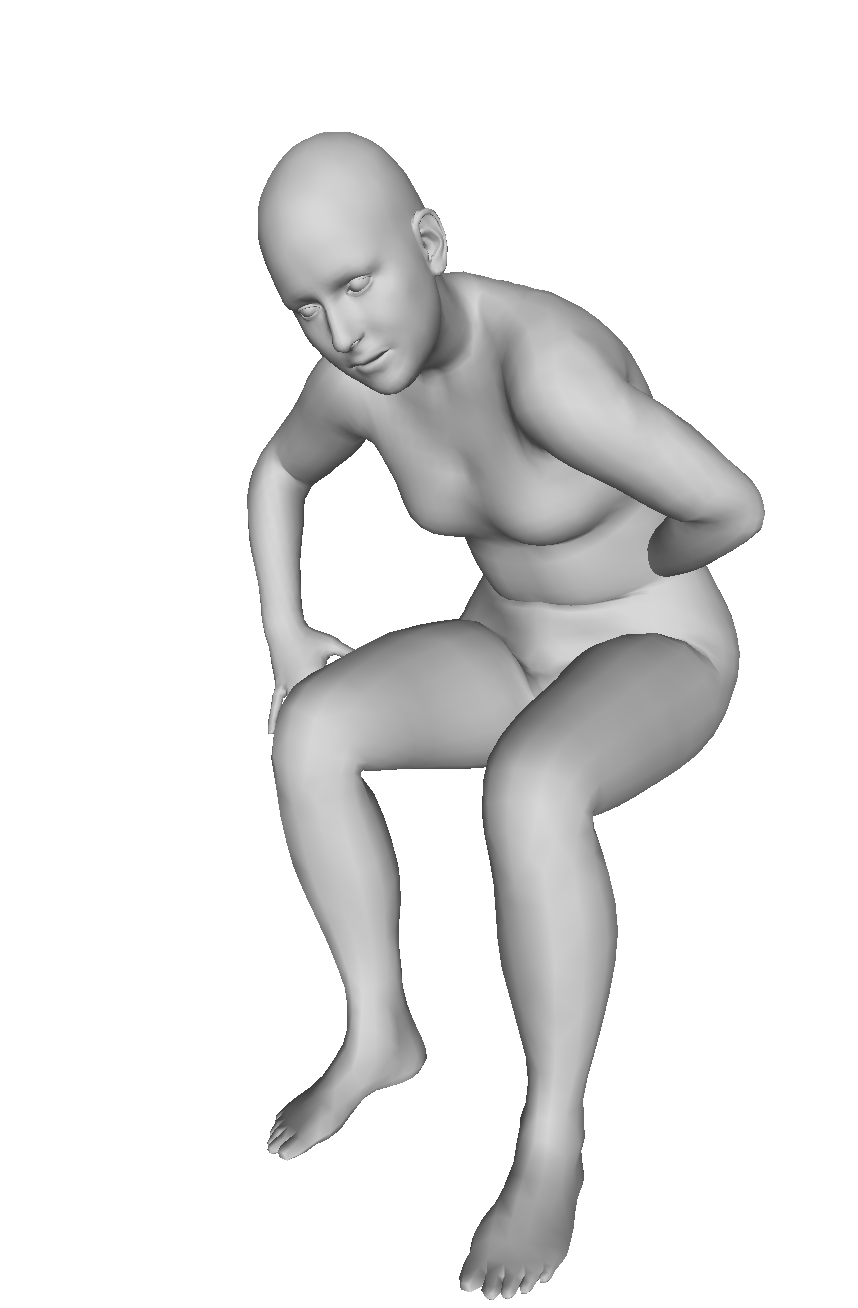} &
        \includegraphics[width=\sz\linewidth]{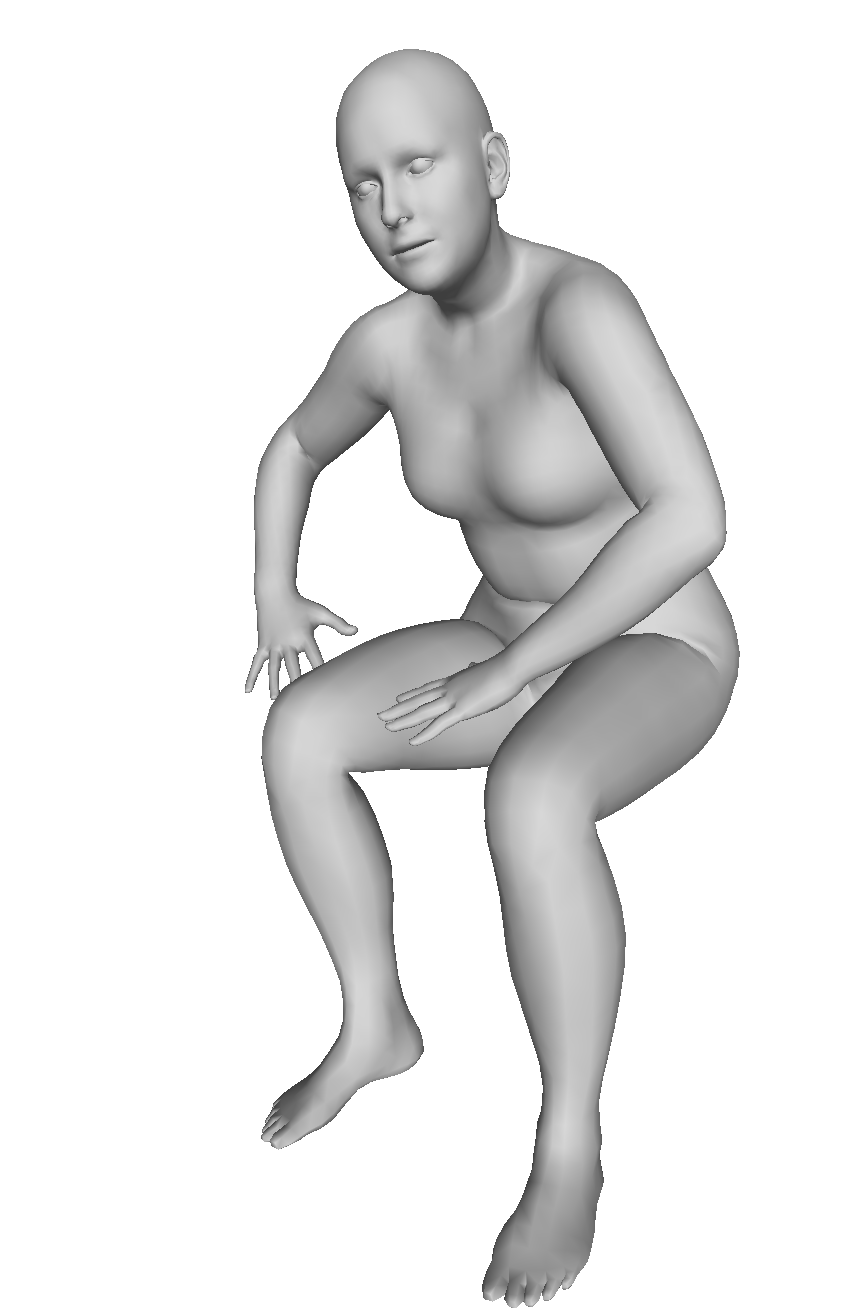} \\
        
    \end{tabular}
    
    \caption{\textbf{Resolving self-interactions.} Comparison of our method and the baseline~\cite{Tzionas:IJCV:2016, pavlakos2019expressive} on the PROX~\cite{PROX:2019} dataset. Our optimization method with COAP can  resolve  highly  ill-posed  self-intersections  such  as  a hand penetrating deeply into the torso.}
    \label{fig:self_pen_imgs}
\end{figure*}

% \begin{figure}
%     \begin{center}
%         \includegraphics[width=0.8\linewidth,height=6cm,draft]{figures/self_pen_imgs.png}
%     \end{center}
%     \caption{\textbf{Self-intersections -- placeholder} PROX vs Baseline~\cite{Tzionas:IJCV:2016} vs Ours}
%     \label{fig:self_pen_imgs}
% \end{figure}

\paragraph{Evaluation}
%To demonstrate the effectiveness of our method, w
We use the PROX dataset~\cite{PROX:2019} to study the effectiveness of our method. This dataset contains invalid 3D human bodies whose body parts intersect with each other. 
From PROX, we sample 100 SMPL bodies by checking the number of self-intersected mesh triangles and compare our method (trained on the MoVi dataset \cite{ghorbani2020movi}) with the commonly used mesh-based method \cite{Tzionas:IJCV:2016, pavlakos2019expressive}\footnote{Code of \url{github.com/vchoutas/torch-mesh-isect}} that penalizes intersected mesh triangles via local distance fields. 

Both methods optimize the input pose parameters with a simple gradient descent until convergence or the maximum of $200$ optimization steps. We quantify the model performance by computing the mean number of self-intersected triangles in the SMPL meshes. %Results shown in 
\figurename~\ref{fig:self_pen_chart} illustrates the convergence curve of both methods over $200$ optimization steps.
Note that our method converges significantly faster and achieves better results compared to the baseline due to the key advantage that our loss term is volume-aware, whereas the baseline imposes the penalty only on the mesh surface. 
%\textbf{(2)} our implicit representation yields spatially continuous gradients, while the baseline produces discontinuous gradients due to the discretization of the space based on colliding mesh triangles.

Qualitative results in \figurename~\ref{fig:self_pen_imgs} illustrate that our method can resolve highly ill-posed self-intersections such as a hand penetrating deeply into the torso. 

\subsection{Resolving Collisions with 3D Environments}\label{subsec:scenepen}
\paragraph{Method}
Our method is also compatible with scene-aware human reconstruction methods \cite{PROX:2019, rempe2021humor, LEMO:Zhang:ICCV:2021}. 
These methods convert raw scans of 3D scenes into SDF grids in order to handle collisions. However, such a process is costly and not always feasible. 
With our representation, one can easily resolve such collisions directly with the raw scans $\mathcal{R} = \{r \in \mathbb{R}^3\}$ by using the following loss term:
% 
% Since our human body model is represented as a parametric implicit function, data points from scene scans $\mathcal{S} \in \mathbb{R}^{N \times 3}$ can be directly use to impose a collision loss term as: 
\begin{equation} \label{eq:collision_loss}
    E_{\text{collision}}(\mathcal{R}) = \frac{1}{|\mathcal{R}|}\sum\nolimits_{r \in \mathcal{R}} \sigma\left(f_\Theta(r|\mathcal{G})\right) \mathbb{I}_{f_\Theta(r|\mathcal{G}) > 0}\,.
\end{equation}

\paragraph{Evaluation} 
We demonstrate this application on the lab-controlled portion of the PROX dataset \cite{PROX:2019}, which has accurate scene SDF grids and SMPL registrations (the PROX Quantitative dataset). To impose the collision loss $E_{\text{collision}}$ \eqref{eq:collision_loss}, we directly samples points from a given 3D scan and shift them along the opposite direction of the scan's surface orientation by a displacement sampled from a normal distribution $\mathcal{N}(0.05, 0.05)$. This collision term is then added to the reconstruction terms from the PROX reconstruction pipeline, including 2D joint reprojection $E_J$, human pose priors $E_P$, and contact $E_C$ loss terms (see \cite{PROX:2019} for more details).
The final reconstruction loss term is defined as:
\begin{equation} \label{eq:full_collision_loss}
    E = E_J + E_P + E_C + E_{\text{collision}}\,,
\end{equation}
which is then optimized with the L-BFGS optimizer \cite{nocedal2006nonlinear} until convergence. We see in \tablename~\ref{tab:collisions_prox} that our method improves the reconstruction accuracy and produces more physically plausible human bodies by reducing collisions with the environment. We also provide the analysis assuming the collision term is derived from a ground truth scene SDF $E^{\text{GT SDF}}_{\text{collision}}$ (third row) for reference. 

We refer the reader to the supplementary video and material for qualitative results and cases where the proposed optimization fails.  

\begin{table}
    \centering
    \scriptsize
    \setlength{\tabcolsep}{4.4pt}
    \newcommand{\ccol}[1]{\textcolor{CadetBlue}{#1}} % custom color
    \begin{tabular}{@{}lccc@{}}
        \toprule
                                                                 & V2V [mm] $\downarrow$ & PJE [mm] $\downarrow$ & Penetration $\downarrow$ \\
        \midrule
        $E_J + E_P + E_C$                                        & 154.26                & 154.39            & 143.52    \\
        $E_J + E_P + E_C + E_{\text{collision}}$                 & \textbf{154.15}       & \textbf{154.34}   & \textbf{100.17}    \\
        \midrule
        $E_J + E_P + E_C + E^{\text{GT SDF}}_{\text{collision}}$ & 154.01  & 154.13     & 46.84    \\
        \bottomrule
    \end{tabular}
    \caption{
    \textbf{Collisions with environment.} 
    Experiment performed on the PROX quantitative dataset \cite{PROX:2019}. 
    We report the mean vertex-to-vertex error (V2V), the mean per-joint error (JPE), and the mean number of penetrated SMPL mesh vertices into the 3D scene geometry (Penetration). We also provide the analysis assuming the collision term is derived from a ground truth scene SDF $E^{\text{GT SDF}}_{\text{collision}}$ (third row) for reference.}
    \label{tab:collisions_prox}
\end{table}
% \begin{figure}
%     \begin{center}
%         \includegraphics[width=\linewidth]{figures/ContactLoss.pdf}
%     \end{center}
%     \caption{\textbf{Contact/Collision with the environment; left PROX[], right Ours.} TODO: fig's taken from some old LEAP experiment.}
%     \label{fig:contact_loss}
% \end{figure}

\subsection{Limitations} \label{subsec:limitations}
\begin{figure}
    % \vspace{-6pt}
    \begin{center}
        \includegraphics[width=1.0\linewidth]{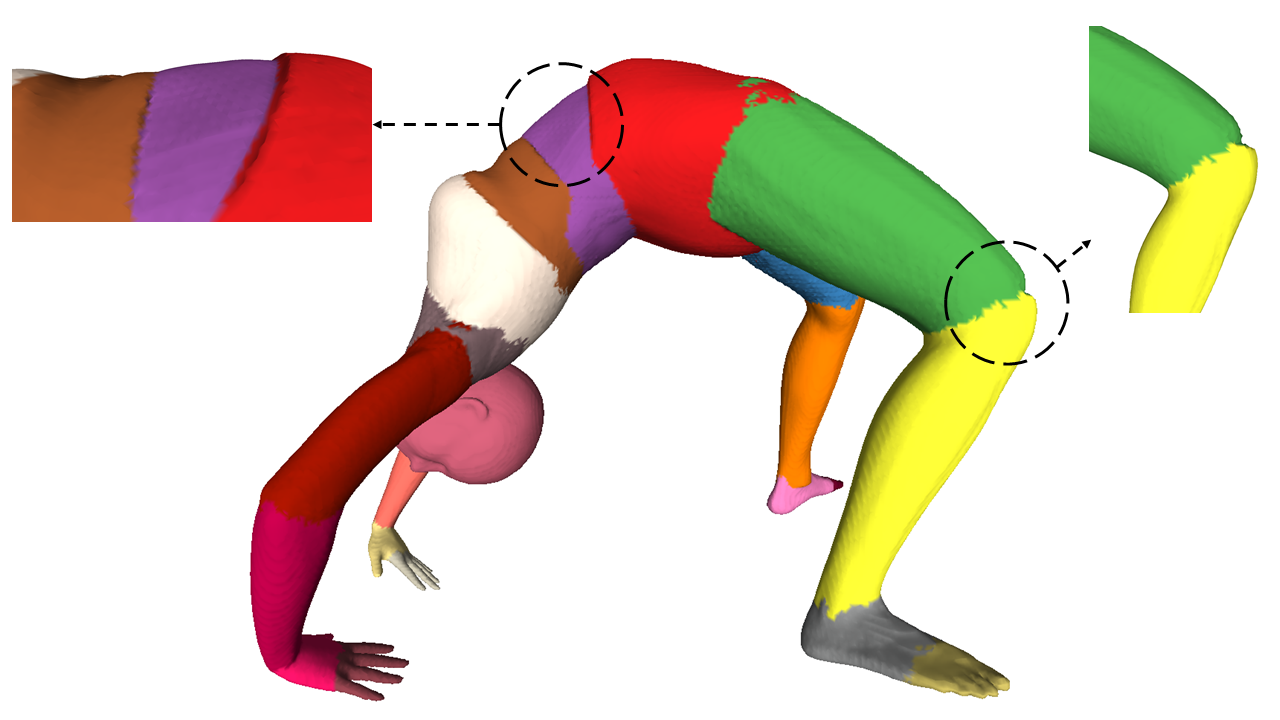}
    \end{center}
    \caption{\textbf{Limitation.} COAP has difficulties modeling smooth transitions between body parts for out-of-distribution poses. The displayed example is a sample from the PosePrior dataset\cite{PosePrior_Akhter:CVPR:2015}.}
    \label{fig:coap_limit_artifacts}
\end{figure}

% \paragraph{Limitations and Failure Cases}
Although COAP performs significantly better than previous state-of-the-art models for neural implicit bodies \cite{chen2021snarf,NGIF,LEAP:CVPR:21} in terms of reconstruction accuracy, sometimes we observe non-smooth connections between body parts (\figurename~\ref{fig:coap_limit_artifacts}) and weak generalization to out-of-distribution extreme body shapes (\eg subject 50002 in Tab.~\ref{tab:sota_single_person} 3rd row) if the model is trained on the small number of diverse identities.
Additionally, the proposed optimization algorithm for resolving self-intersections sometimes can produce less realistic human pose due to the lack of additional terms that incentivize pose naturalness. 
We believe that the inference time of COAP ($75$ms for 10k points) could be improved as it is currently slower compared to the generalizable human bodies LEAP ($35$ms) and Neural-GIF ($22$ms).
Therefore, exploring even more powerful neural representations and optimization pipelines is an interesting direction for future work.
%
% Please see the Supp.~Mat.~for failure cases. 

\section{Conclusion and Future Work}
Neural implicit representations for human body modeling are a rising research topic.
Existing state-of-the-art models have difficulties generalizing to unseen poses and shapes.
In this work, we propose COAP, a novel compositional neural occupancy representation, that drastically improves the robustness and the generalization to challenging motions.
We decompose the geometry of a full body into local body parts and learn per-part occupancy representations by leveraging the geometric constraints facilitated by the prior knowledge of human body shape. 
Such part-aware representation enables efficient untangling of challenging self-intersected human bodies and collision detection with other objects. 

\myparagraph{Future Work.}
As future work we consider modeling clothing for our neural implicit body model, deploying COAP into 3D human body estimators (\eg~\cite{kolotouros2019learning, kocabas2020vibe, ROMP}) to enforce collision-free predictions during the neural network training, as well as addressing current weaknesses such as generalization to extreme out-of-distribution body shapes and small visible artifacts between body parts for some poses.  

% Although our approach performs significantly better than previous state-of-the-art models in terms of reconstruction accuracy, we sometimes can still observe small artifacts between different body parts.
% %
% Therefore, as future work we consider exploring even more powerful neural representations.
% %
% Other ideas for future work include modeling clothing for our neural implicit body model and the integration of COAP into 3D human body estimators (\eg~\cite{kolotouros2019learning, kocabas2020vibe, ROMP}) to enforce collision-free predictions during the neural network training process.

\myparagraph{Acknowledgments.}
We thank Shaofei Wang and Yan Zhang for proofreading and Garvita Tiwari for the help with one of the baselines. 
S. T. and M. M. acknowledge the SNF grant 200021\_204840.

\myparagraph{Disclaimer.}
The project was fully completed at ETH Z\"{u}rich. It was not funded by Meta, nor has it been conducted at Meta.

%%%%%%%%% REFERENCES
{
    % \clearpage
    \small
    \bibliographystyle{ieee_fullname}
    \bibliography{macros,main}
}

% --- supplementary material
\clearpage
\appendix

% --- PDF will be split by an editor (e.g. macOS preview), so need to restart from page 1
\setcounter{page}{1}

% --- repeat the title (AT: haven't found a more elegant way to do this...)
\twocolumn[
    \centering
    \Large
    \textbf{COAP: Compositional Articulated Occupancy of People} \\
    \vspace{0.5em}Supplementary Material \\
    \vspace{1.0em}
] %< twocolumn
\appendix
\setcounter{page}{1}
\setcounter{table}{0}
\setcounter{figure}{0}
\setcounter{equation}{0}
\renewcommand{\thetable}{\thesection.\arabic{table}}
\renewcommand{\thefigure}{\thesection.\arabic{figure}}
\renewcommand{\theequation}{\thesection.\arabic{equation}}

In this supplementary document, we provide additional implementation details (Sec.~\ref{sec:sup:impl}) and qualitative and quantitative results (Sec.~\ref{sec:sup:results}).

\section{Implementation Details}\label{sec:sup:impl}
\begin{figure*}[t!]
\scriptsize
\begin{center}
    \includegraphics[width=\textwidth]{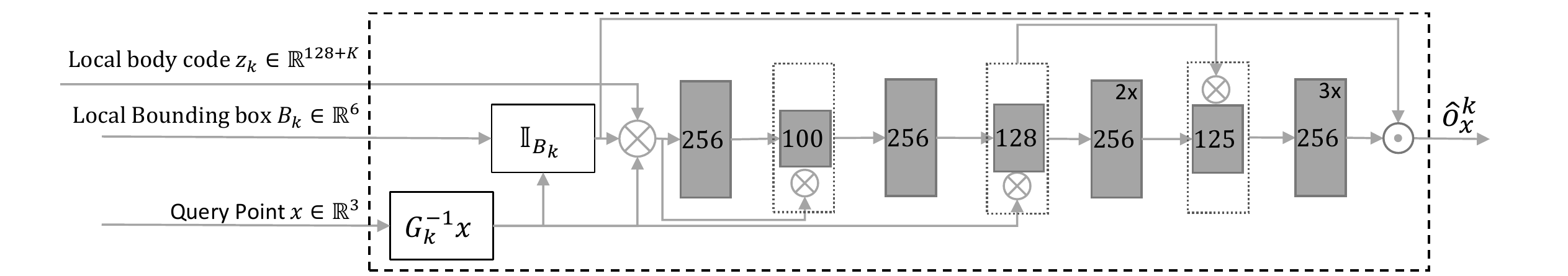}
\end{center}
\caption{\textbf{Architecture of the local MLP decoder.}
The input parameters correspond to the $k$th articulated body part. 
Deterministic differentiable blocks are shown in black,  
fully-connected layers are shown in gray. 
the number inside each block denotes the dimensionality of the input feature vector,
the number in the top-right corner denotes layer repetition, 
the operator $\otimes$ denotes feature concatenation, 
the operator $\odot$ denotes multiplication, 
$\mathbb{I}_{B_k}$ is an indicator function returning the value $1$ if the local query is inside the bounding box $B_k$ or $0$ otherwise.
All fully-connected layers are activated by Softplus with beta of $100$ and a threshold of $20$. 
% The attention block multiplies the initially predicted occupancy value and the binary value that indicates whether the local query is inside the $k$ bounding box. 
The final output of the decoder is the occupancy prediction $\hat{o}_x^k$ for the $k$th articulated part.
}
\label{fig:sup:mlp_deocder}
\end{figure*}

\paragraph{Network Architectures}
The PointNet encoder in Sec.~\ref{sec:representation} is implemented as an eight-layer perceptron network interleaved with ReLU activations and skip connections as in the previous work~\cite{LEAP:CVPR:21}. 
The shared MLP occupancy decoder is illustrated in \figurename~\ref{fig:sup:mlp_deocder}. 

\paragraph{Sampling Strategy in the Local Shape Decomposition (Sec.~\ref{sec:shape_encoder})}
Each local articulated part in \figurename~\ref{fig:overview} is temporarily represented as a point cloud by sampling points on the mesh surface. 
Each point is sampled by first selecting a mesh face with probability proportional to the face area and then randomly sampling barycentric coordinates in order to calculate a point on the selected face. 
To further balance the overlap among local articulated body parts, the $k$th point cloud allocates one half of its capacity to encode the central component corresponding to bone $G_k$, whereas the other half covers the whole local articulated body part region. 
This design guides the neural networks to properly learn localized occupancy fields, where the largest part is reserved to represent the core bone component, while fewer samples for the non-central parts encourage smooth interpolation between connected occupancy fields. 

In all experiments, we used a total of $1000$ samples per body part which are encoded as local body codes $z_k$ with 128 dimensions. 

\section{Additional Results and Experiment Details}\label{sec:sup:results}
\begin{figure}
    \scriptsize
    \setlength{\tabcolsep}{0.0mm} %0
    \newcommand{\sz}{0.25}  % 0.125
    \begin{tabular}{cccc}
        Neural-GIF~\cite{NGIF}   & LEAP~\cite{LEAP:CVPR:21}   & \textbf{COAP} & GT  \\
        \includegraphics[width=\sz\linewidth]{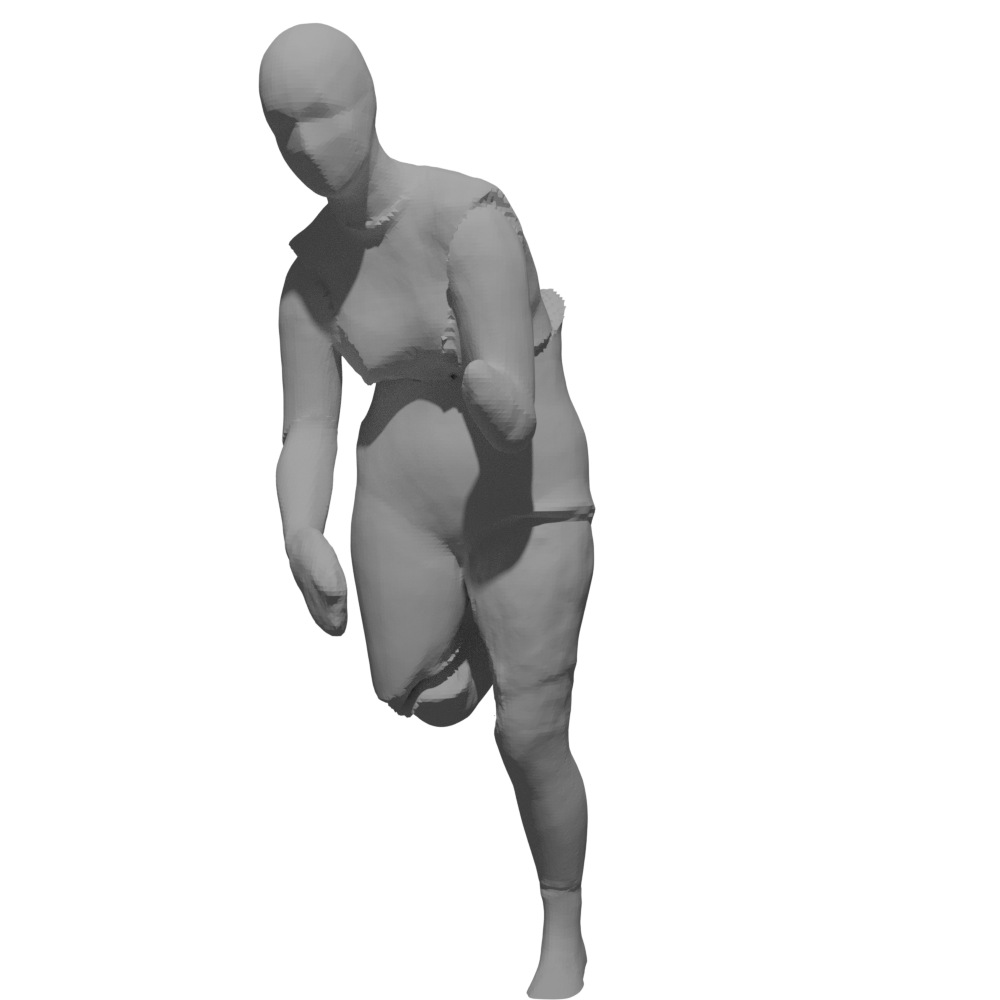} & \includegraphics[width=\sz\linewidth]{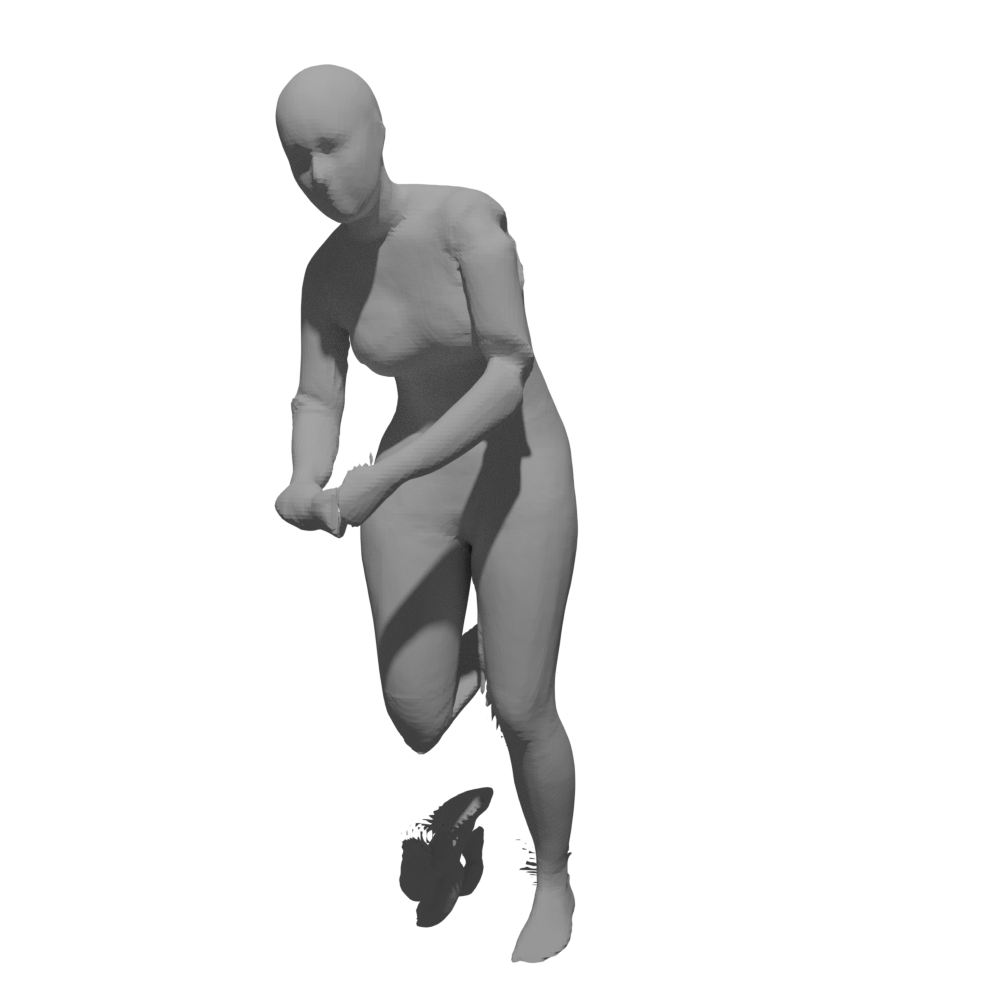} & \includegraphics[width=\sz\linewidth]{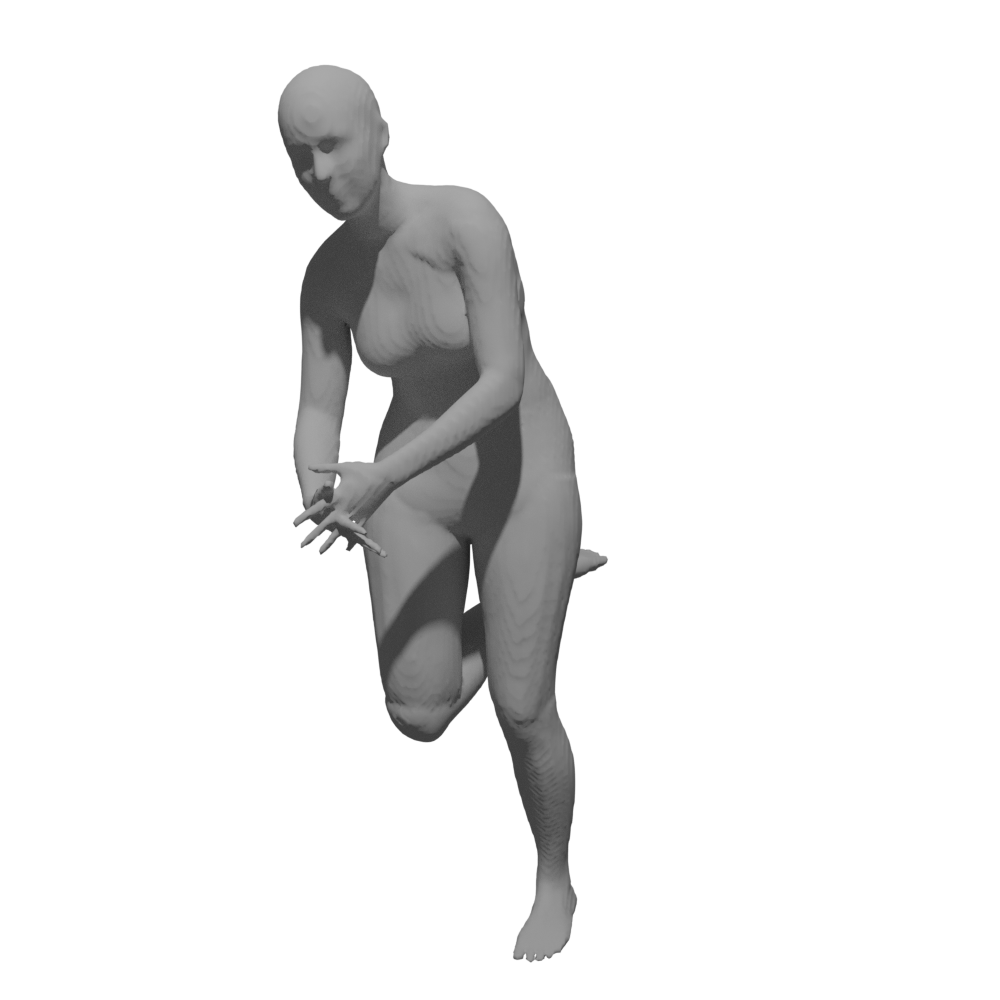}& \includegraphics[width=\sz\linewidth]{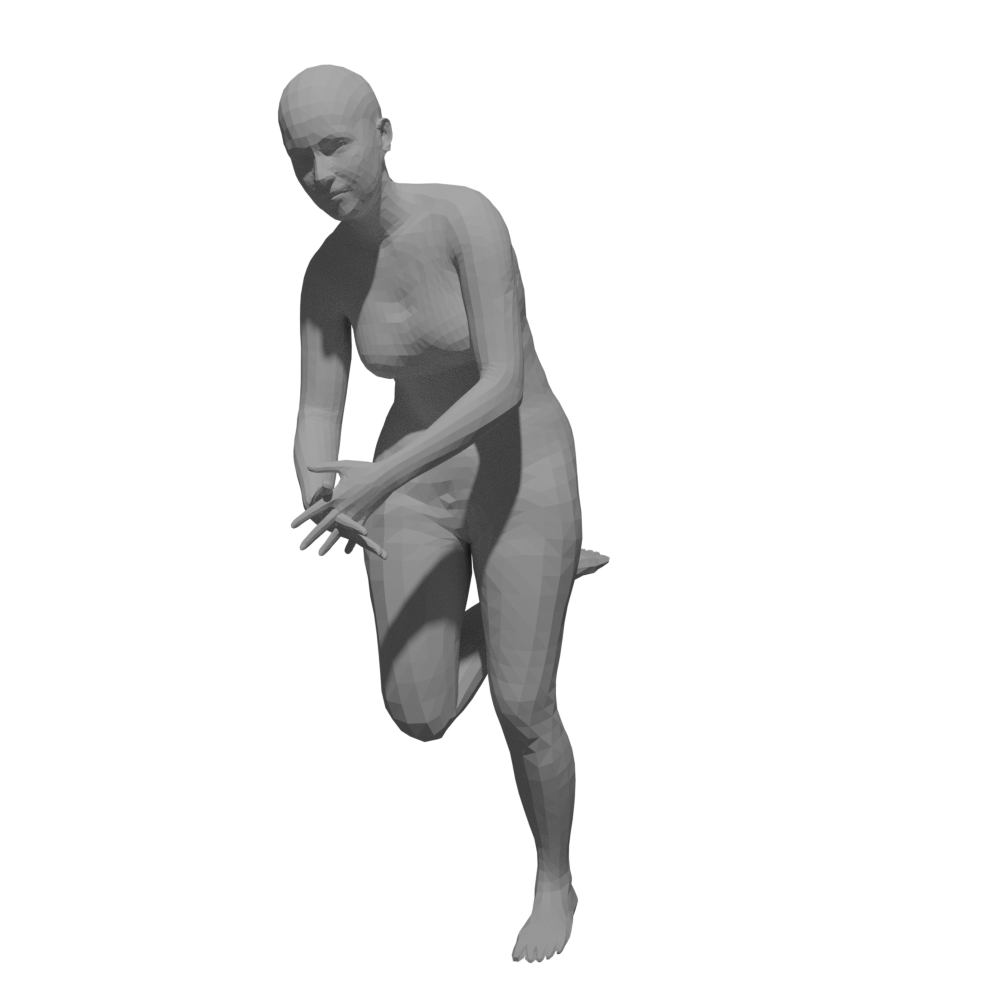}\\
        
        \includegraphics[width=\sz\linewidth]{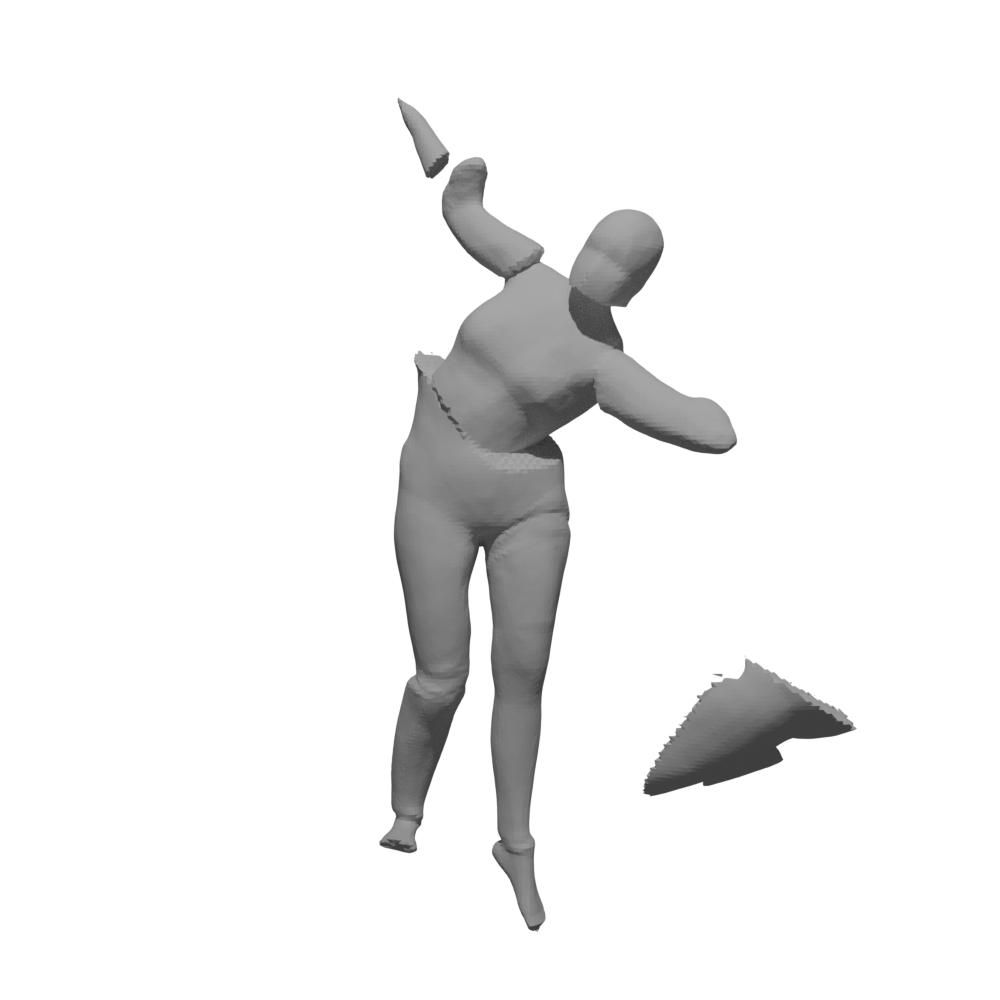} & \includegraphics[width=\sz\linewidth]{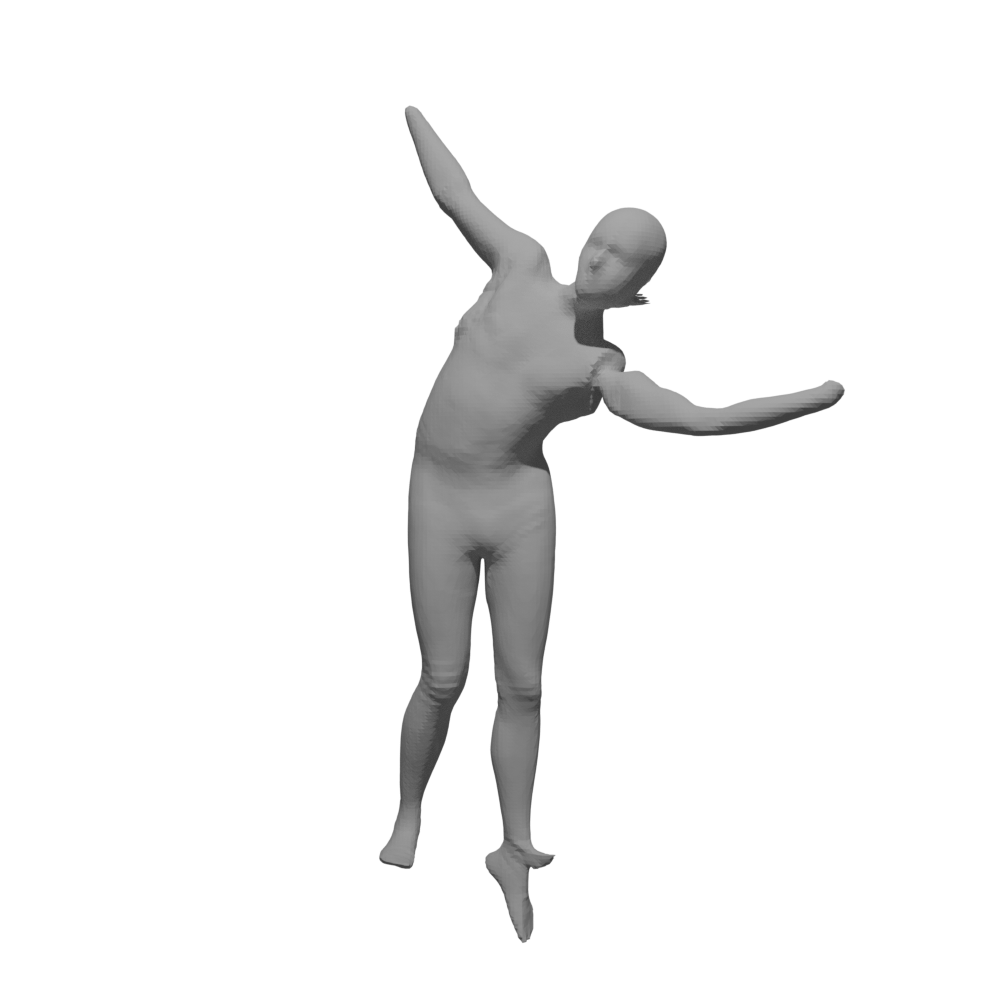} & \includegraphics[width=\sz\linewidth]{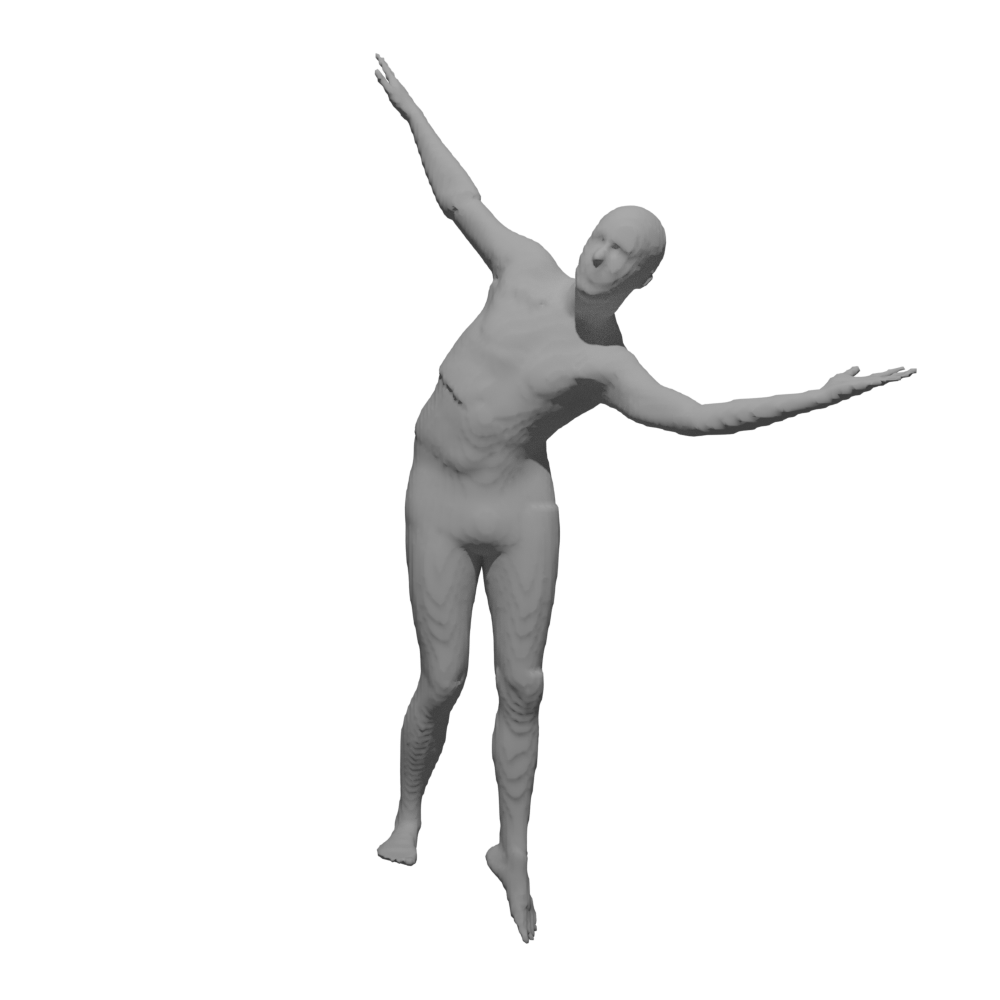}& \includegraphics[width=\sz\linewidth]{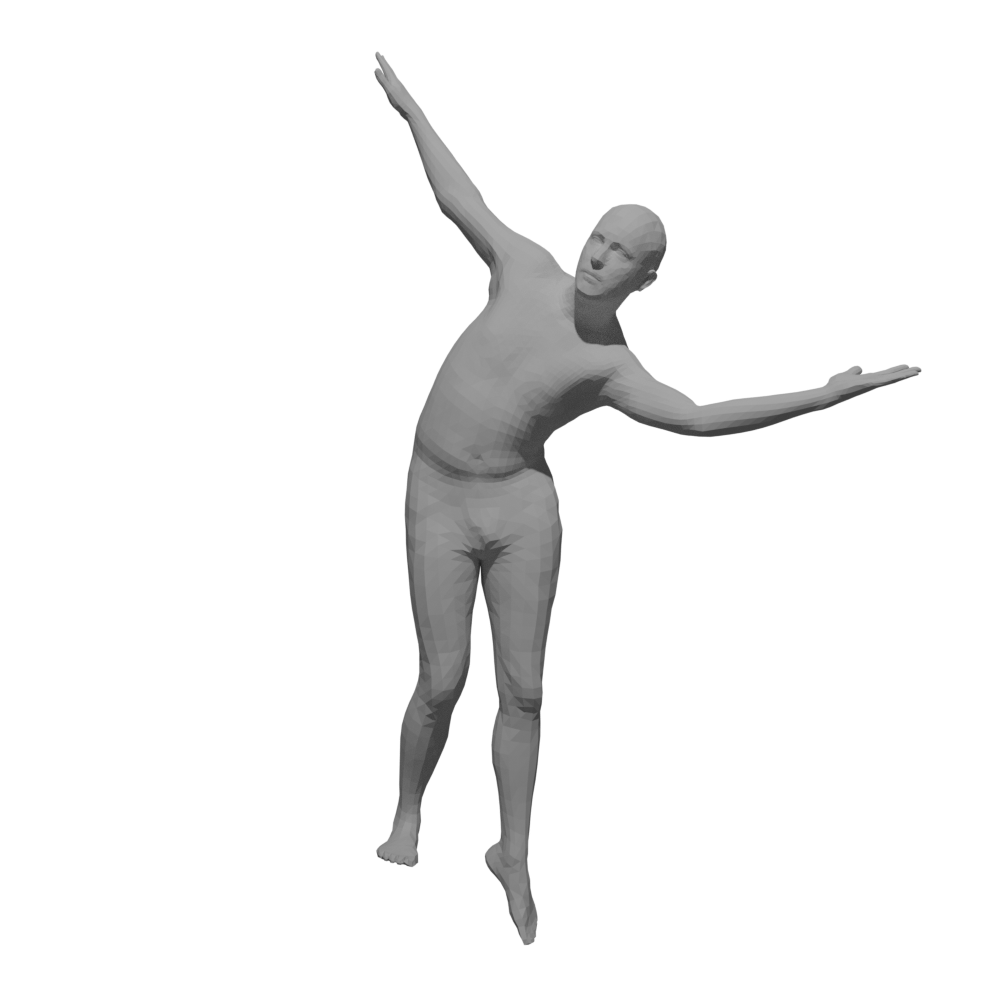}\\
        
        \includegraphics[width=\sz\linewidth]{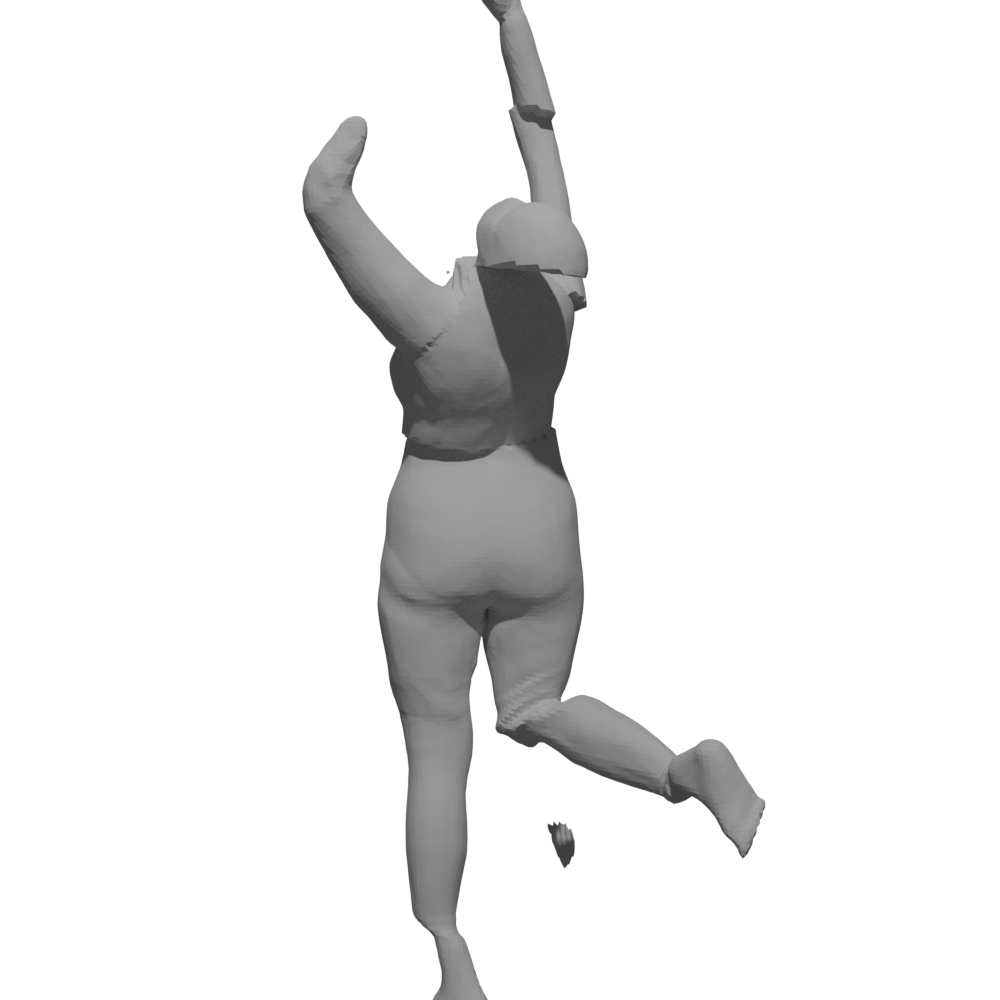} & \includegraphics[width=\sz\linewidth]{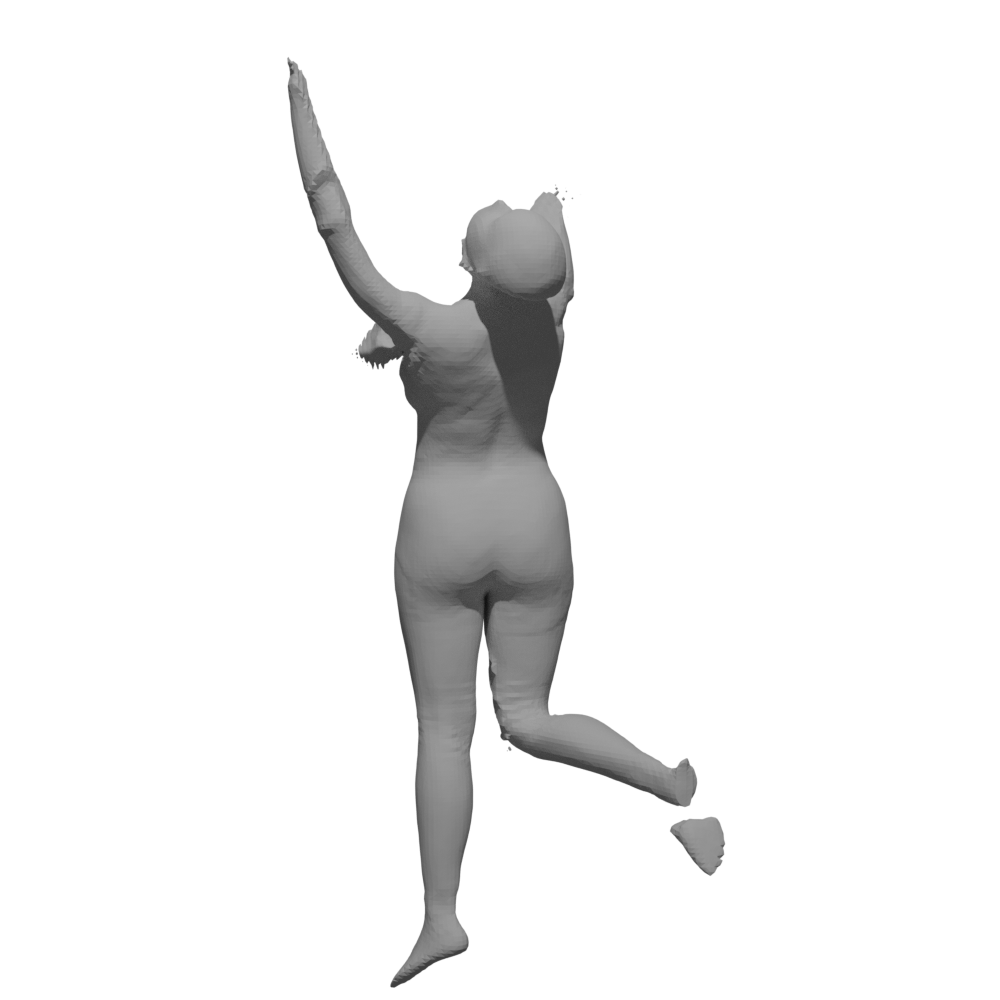} & \includegraphics[width=\sz\linewidth]{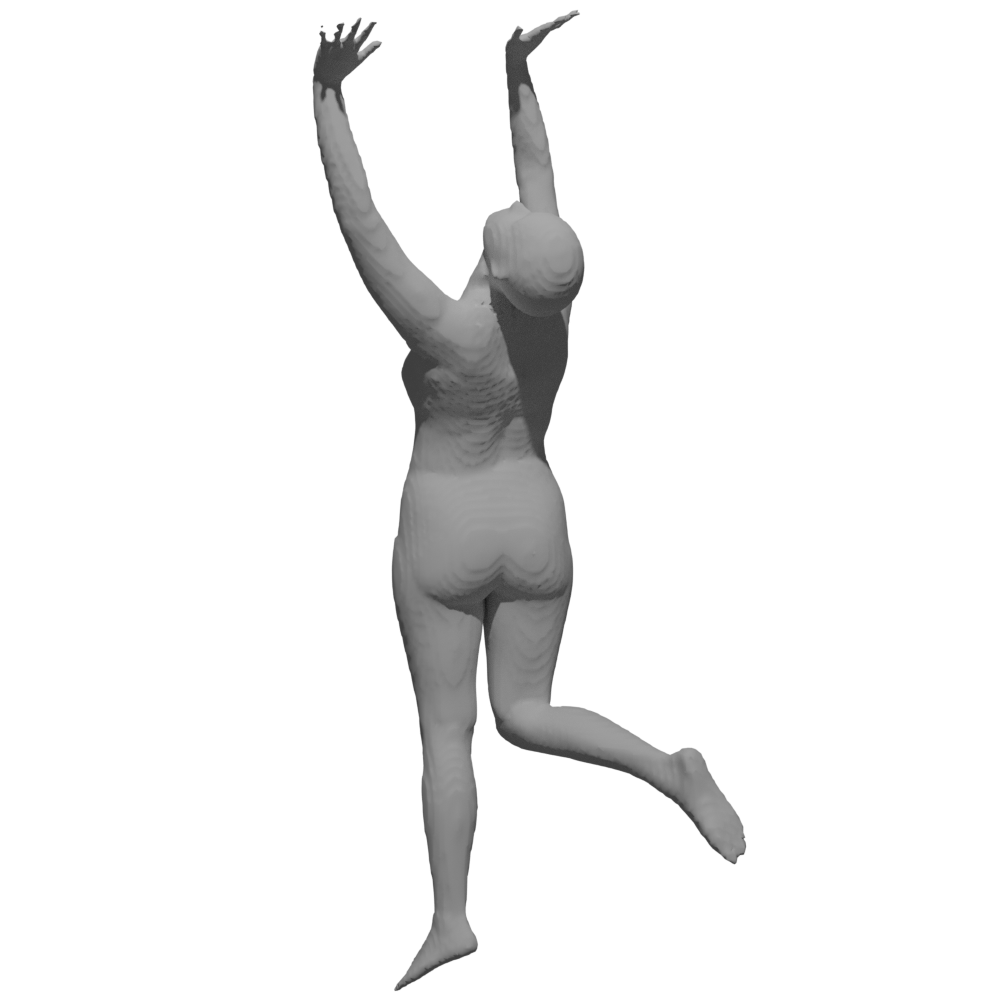}& \includegraphics[width=\sz\linewidth]{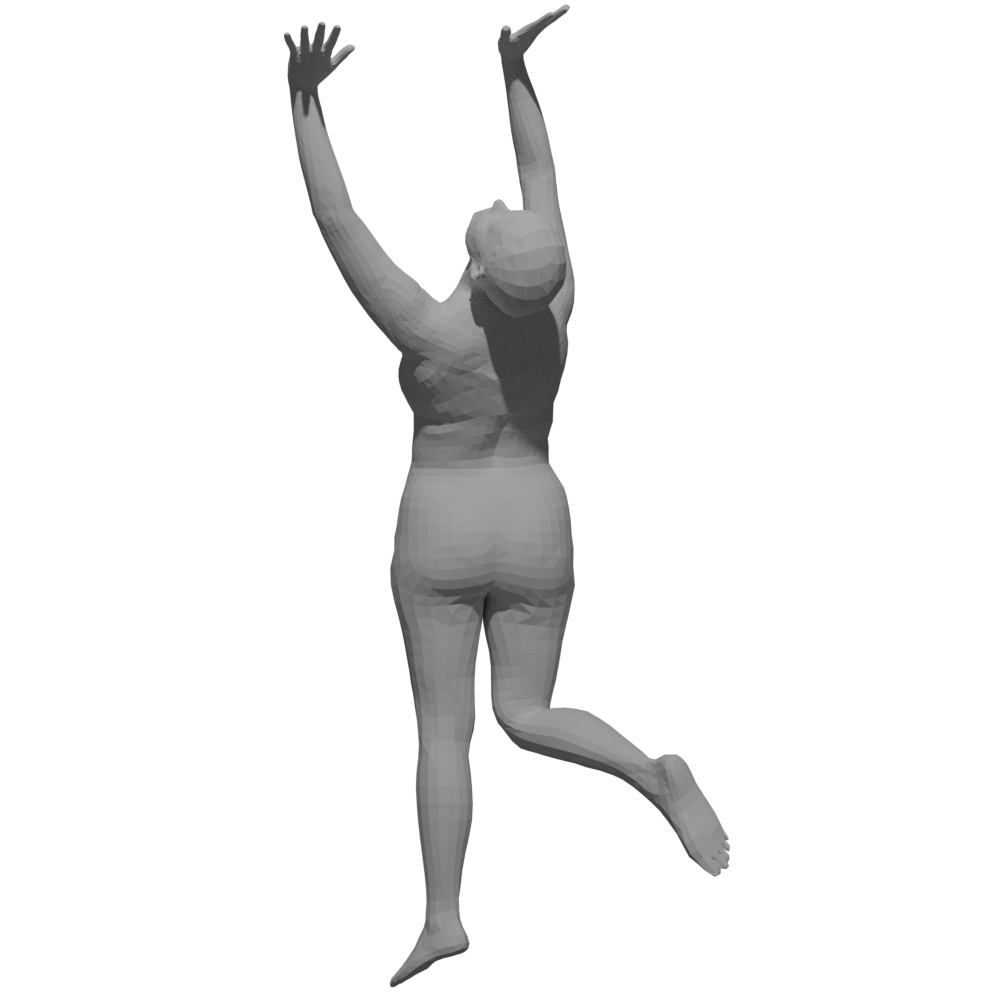}\\

        \includegraphics[width=\sz\linewidth]{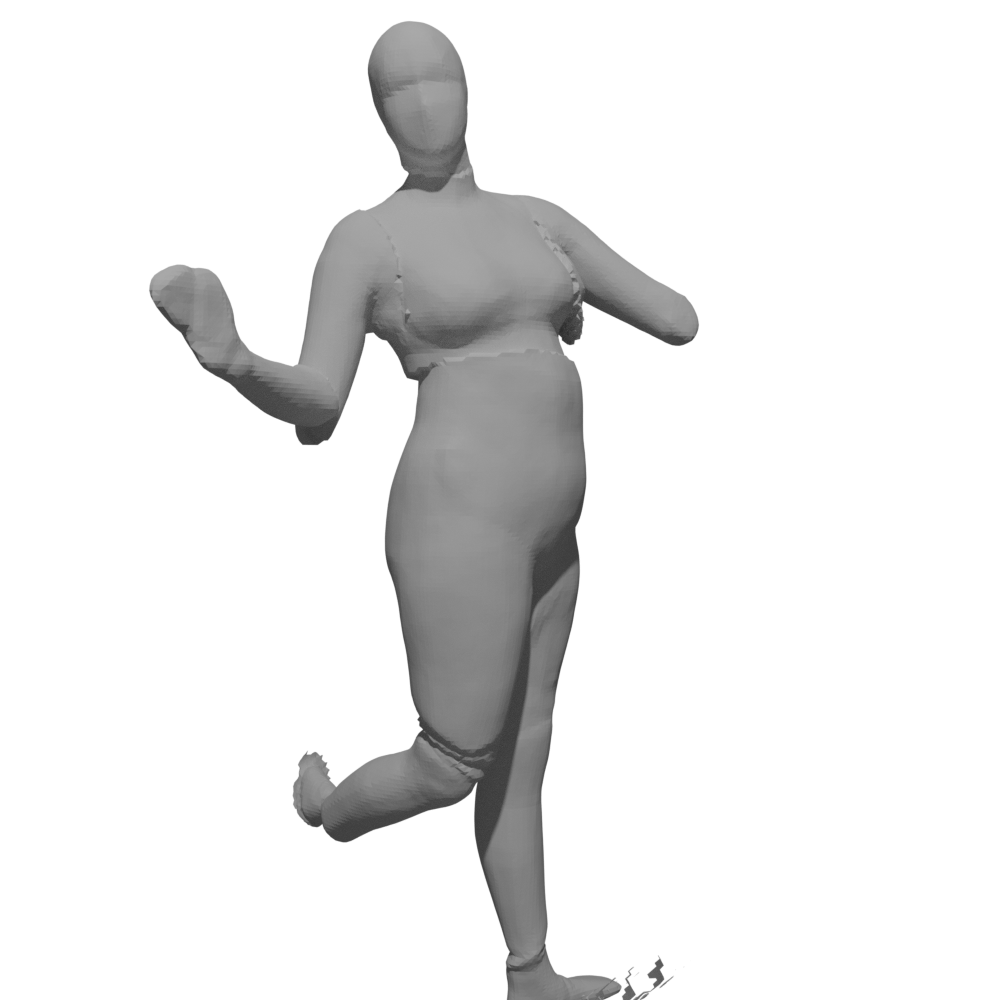} & \includegraphics[width=\sz\linewidth]{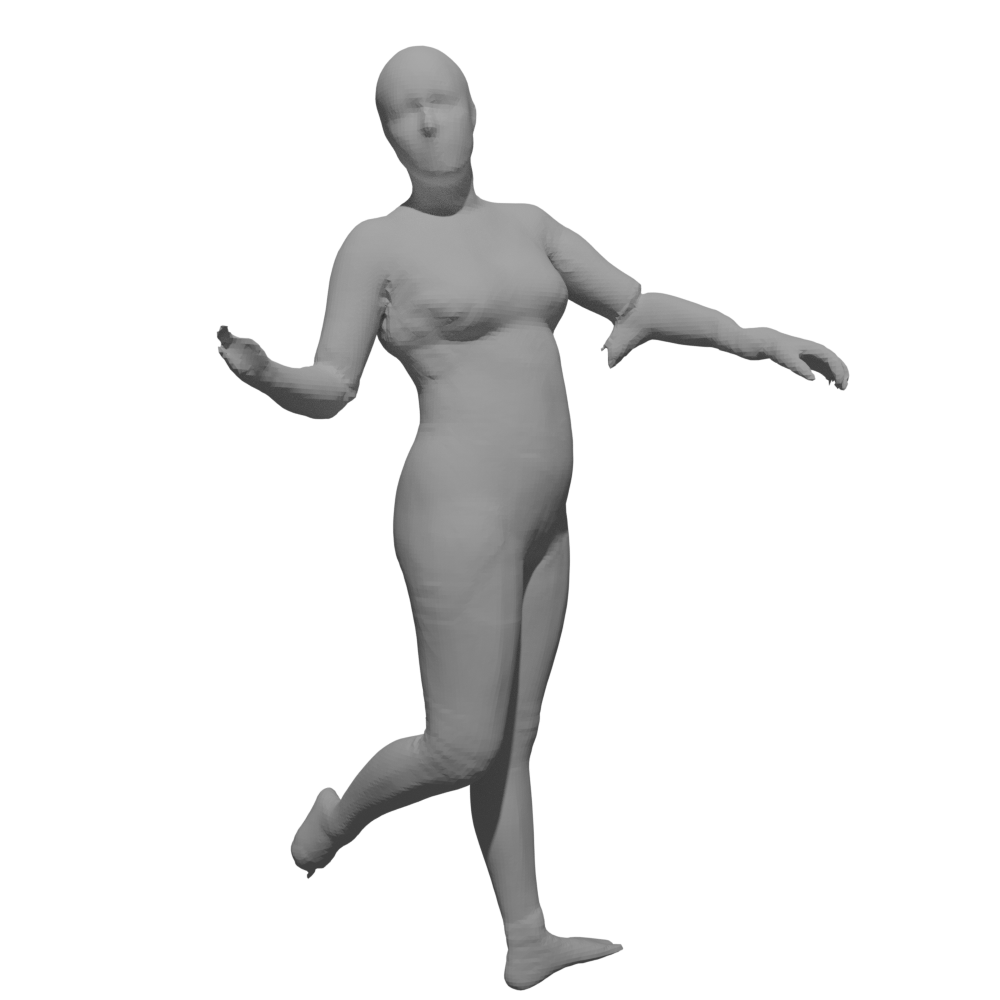} & \includegraphics[width=\sz\linewidth]{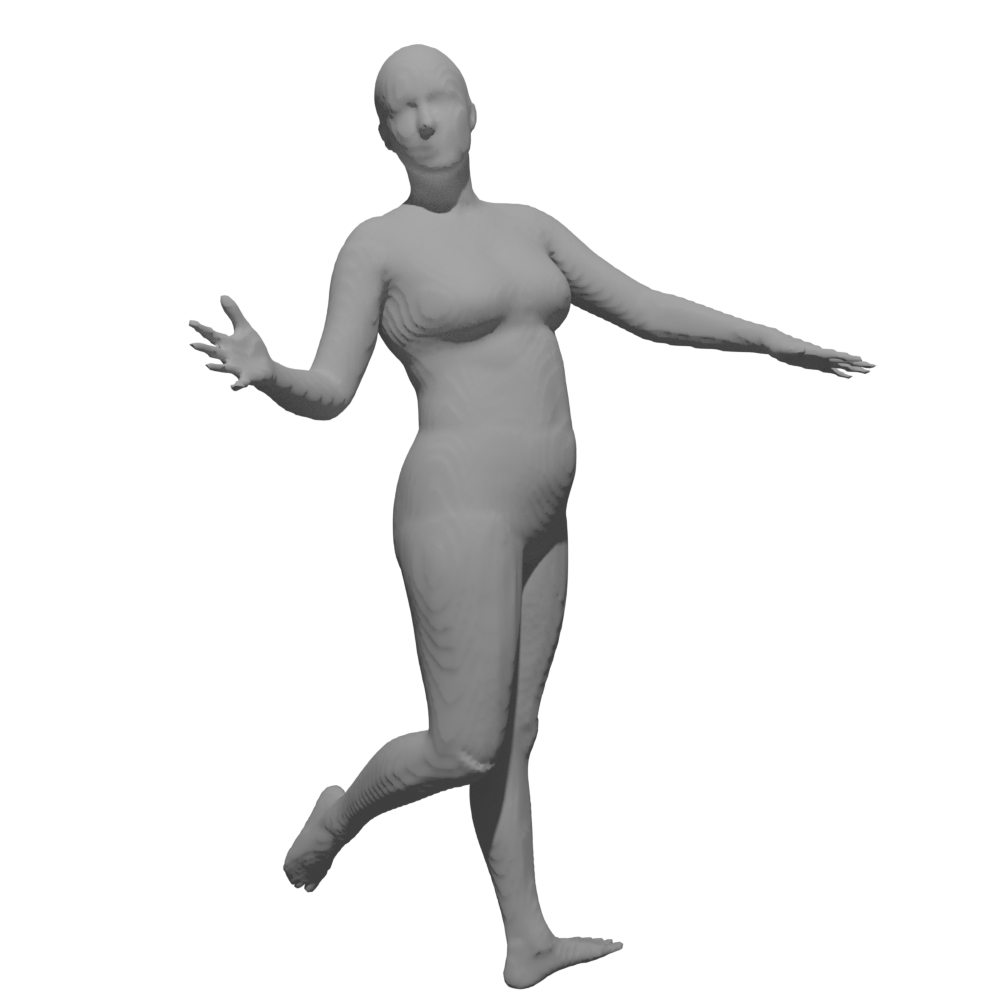}& \includegraphics[width=\sz\linewidth]{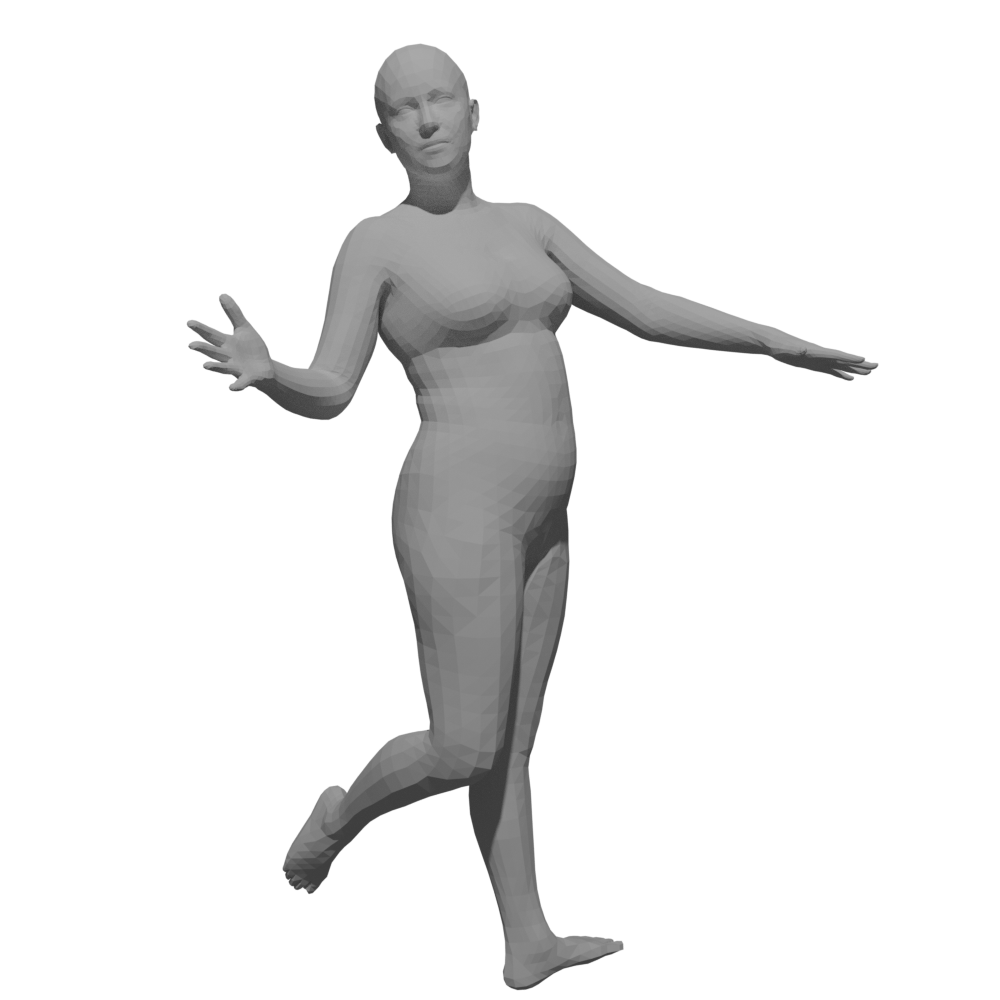}\\

        \includegraphics[width=\sz\linewidth]{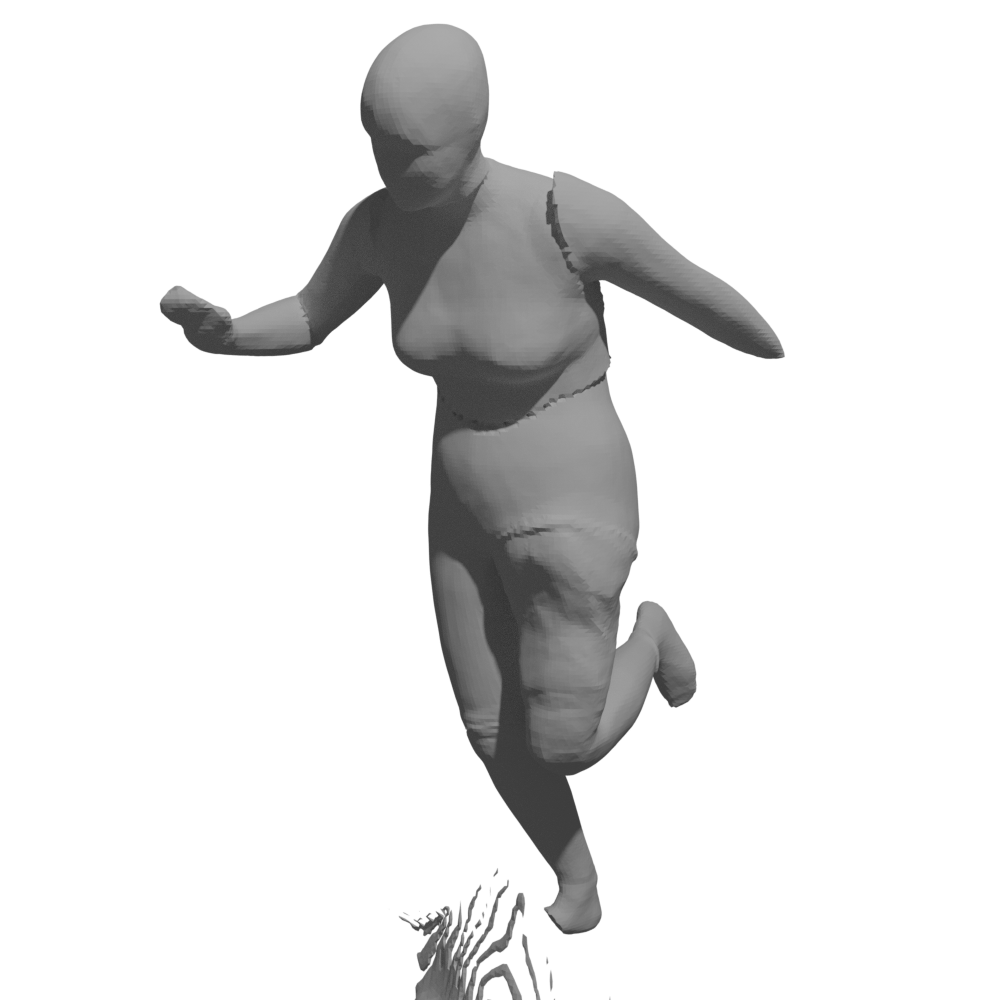} & \includegraphics[width=\sz\linewidth]{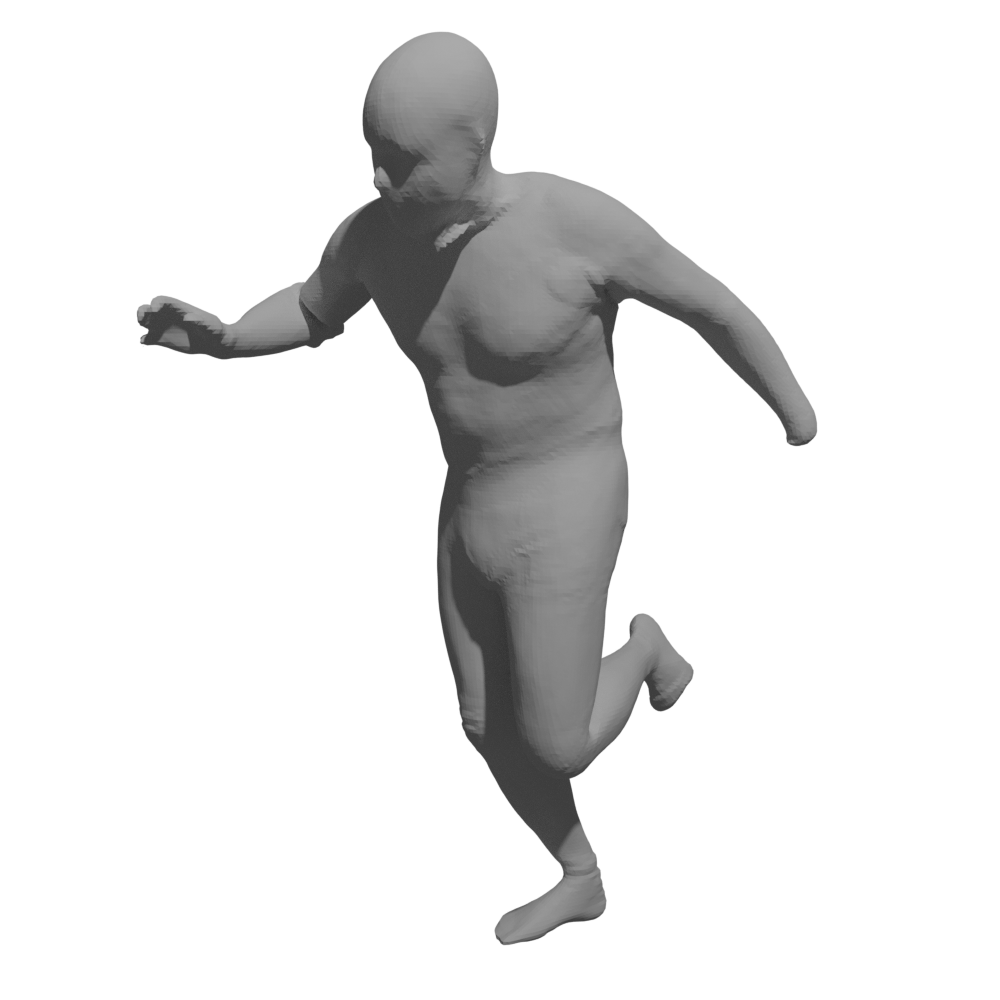} & \includegraphics[width=\sz\linewidth]{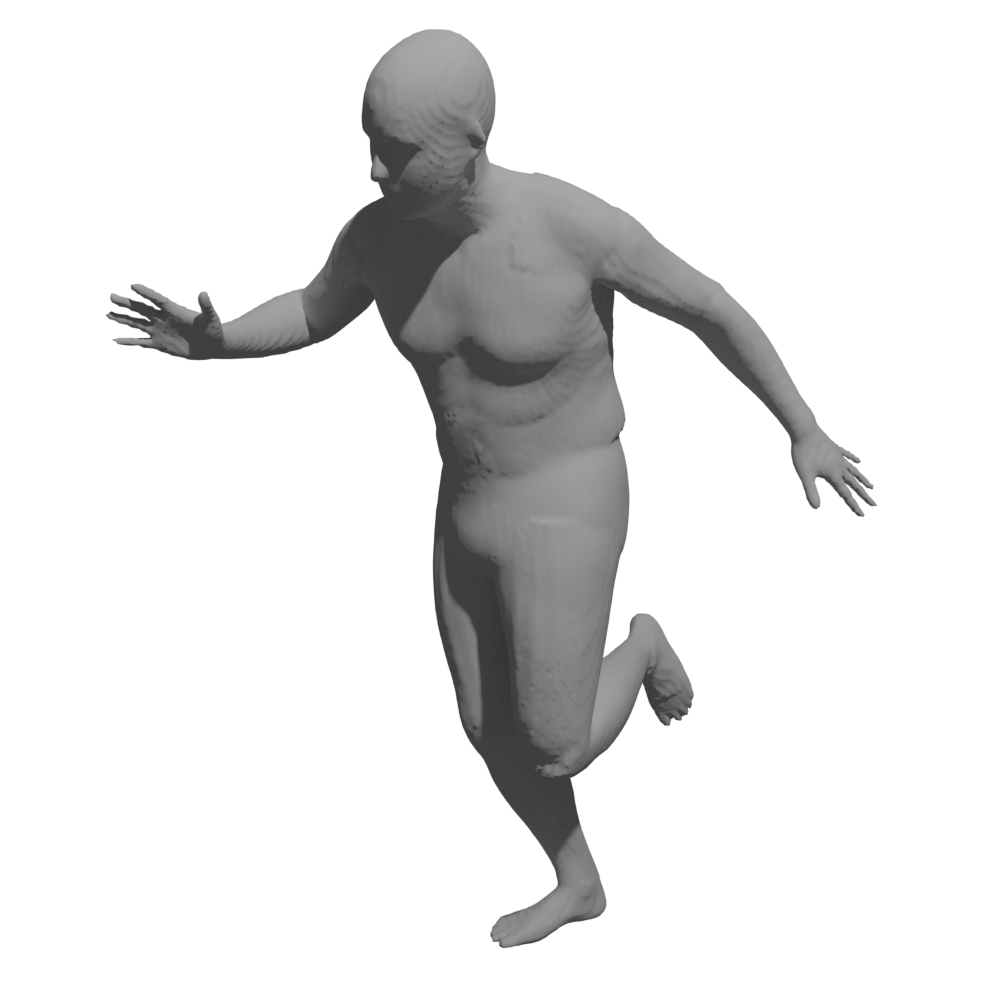}& \includegraphics[width=\sz\linewidth]{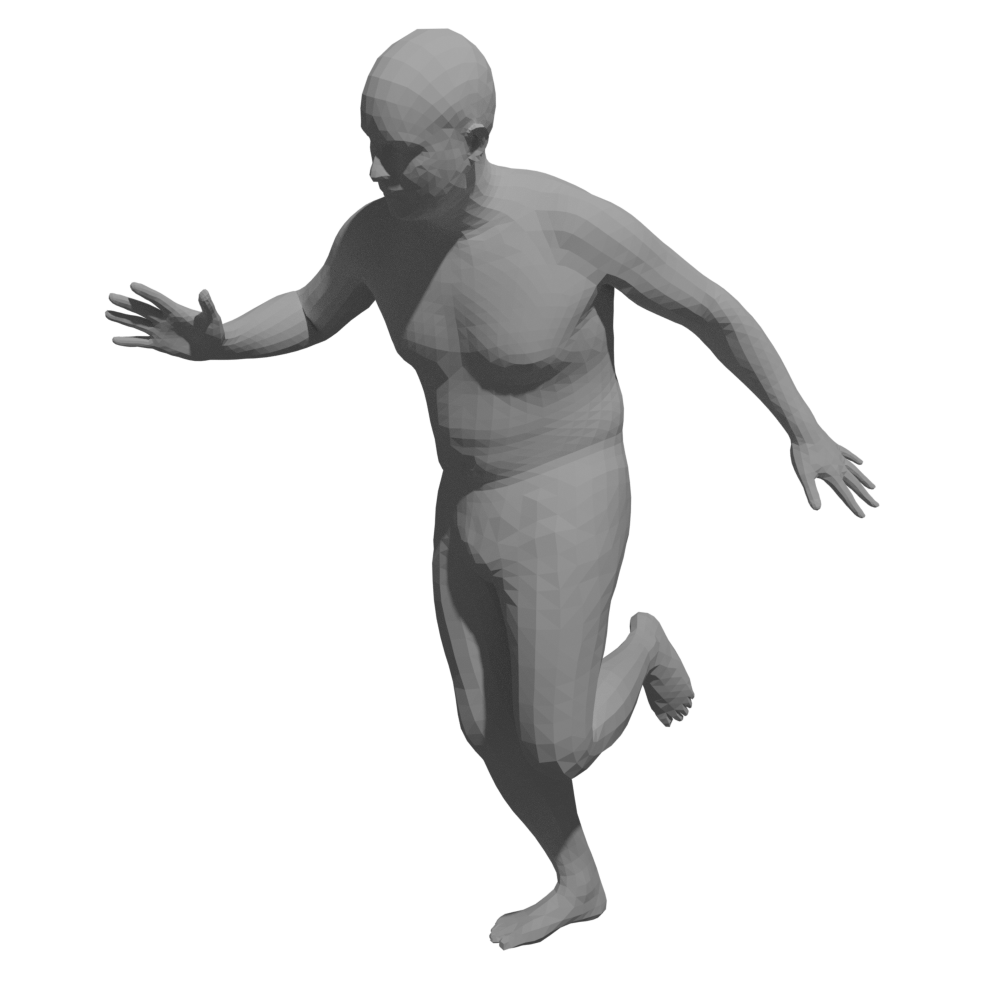}\\

        \includegraphics[width=\sz\linewidth]{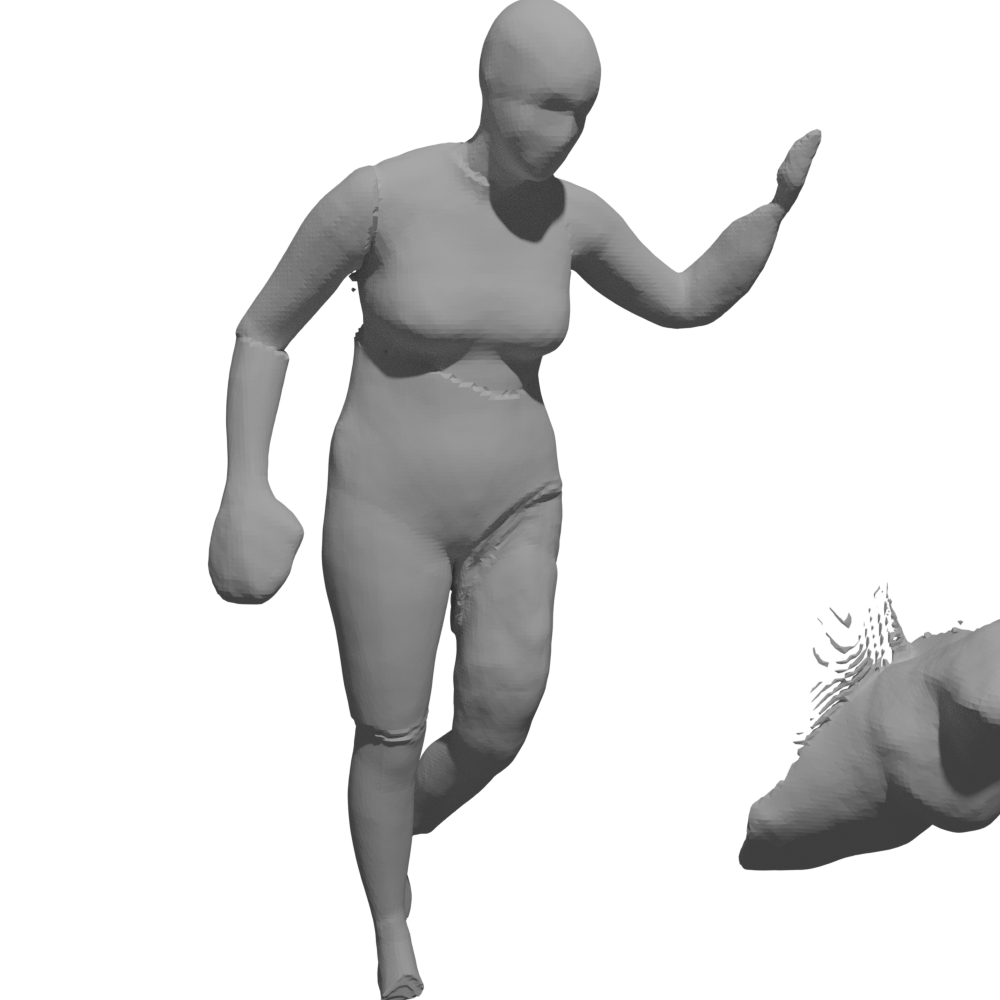} & \includegraphics[width=\sz\linewidth]{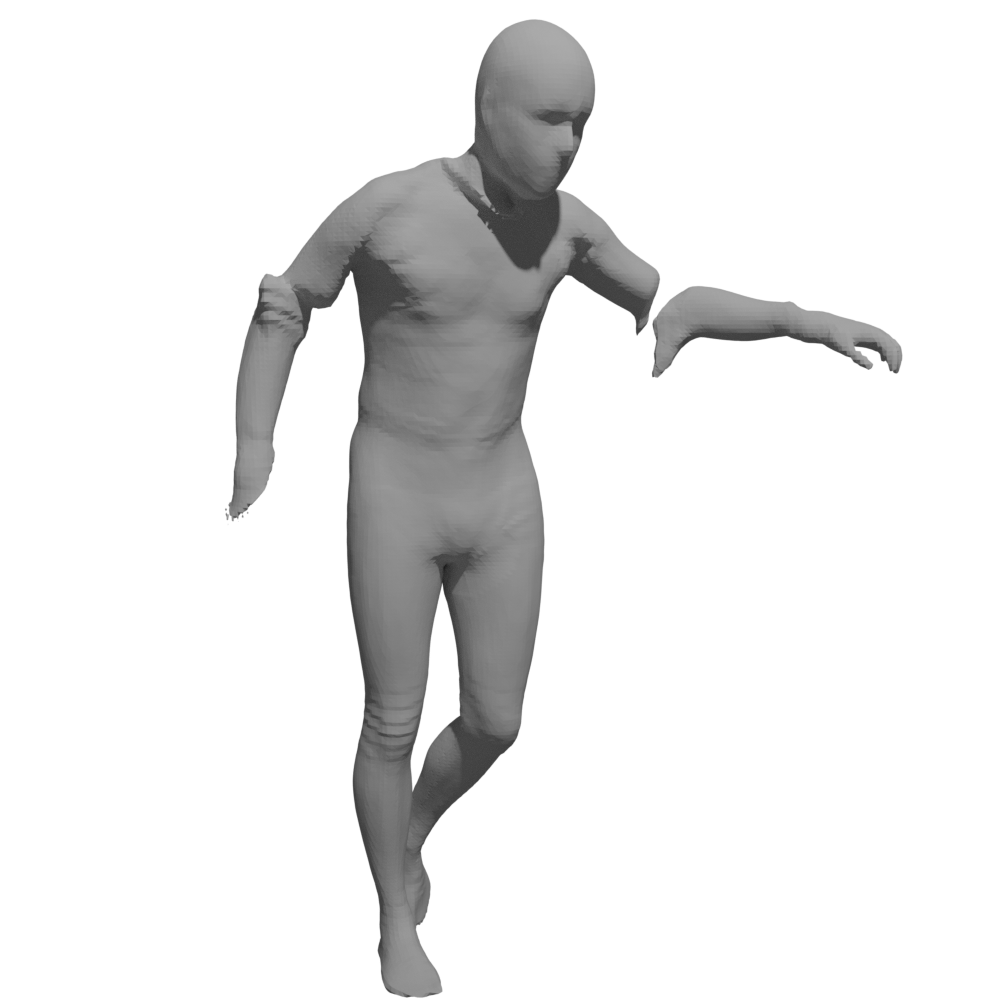} & \includegraphics[width=\sz\linewidth]{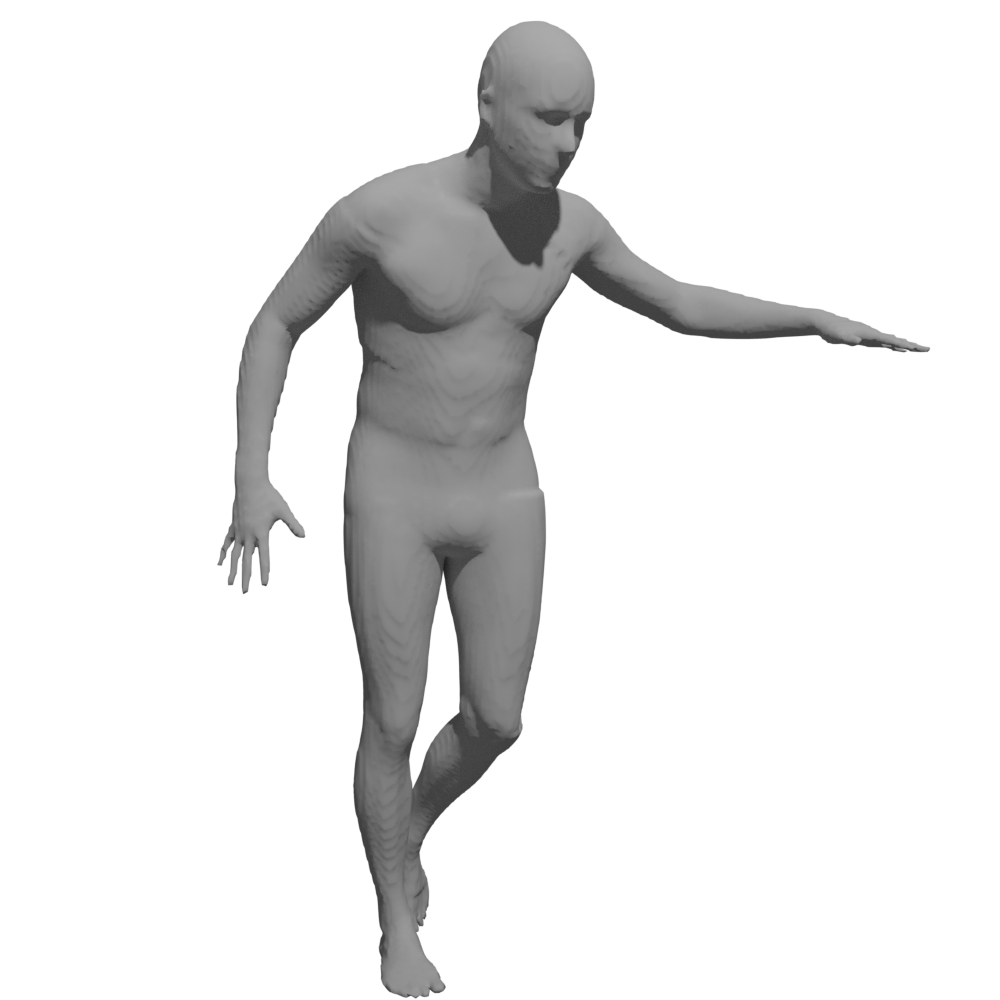}& \includegraphics[width=\sz\linewidth]{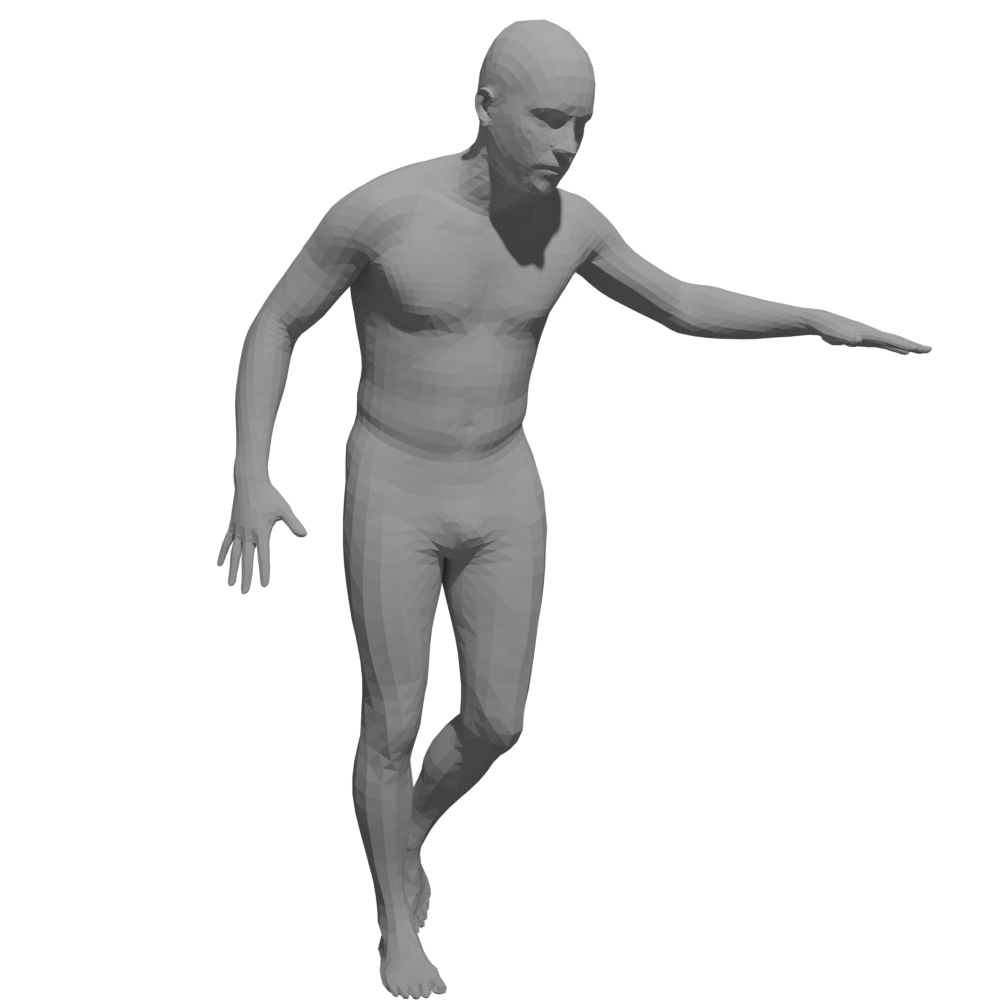}\\

    \end{tabular}
    
    \caption{\textbf{Generalization to unseen humans.}
    Comparison of our model with LEAP~\cite{LEAP:CVPR:21} and Neural-GIF~\cite{NGIF} for the identities of the DFaust~\cite{dfaust:CVPR:2017} and the PosePrior~\cite{PosePrior_Akhter:CVPR:2015} datasets performing challenging novel poses from the PosePrior dataset.
    These qualitative results supplement results displayed in \figurename~\ref{fig:teaser} and \tablename~\ref{tab:sota_unseen_ppl}.}
    \label{fig:sup:sota_unseen_ppl}
\end{figure}

\begin{table*}
\centering
%\scriptsize
\footnotesize
\setlength{\tabcolsep}{3.4pt}

\begin{tabular}{@{}l|ccccc|ccccc@{}}
\toprule
\multicolumn{1}{l}{}        & \multicolumn{5}{c}{Female Subjects}  & \multicolumn{5}{c}{Male Subjects}      \\

Method                      & 50004             & 50020             & 50021             & 50022             & 50025             & 50002             & 50007             & 50009             & 50026             & 50027             \\
\midrule
NASA \cite{chen2021snarf}   & 77.75/77.68       & 55.93/80.20       & 90.99/78.13       & 90.87/77.86       & 71.20/78.64       & 68.14/74.82       & 67.57/71.82       & 44.84/74.32       & 87.44/77.47       & 48.84/79.30       \\
% Neural-GIF \cite{NGIF}    & xx.xx/xx.xx       & xx.xx/xx.xx       & xx.xx/xx.xx       & xx.xx/xx.xx       & xx.xx/xx.xx       & xx.xx/xx.xx       & xx.xx/xx.xx       & xx.xx/xx.xx       & xx.xx/xx.xx       & xx.xx/xx.xx       \\
LEAP \cite{LEAP:CVPR:21}    & 88.53/67.05       & 90.42/77.84       & 89.84/76.15       & 88.18/64.79       & 91.33/77.09       & 74.67/35.31       & 83.65/53.83       & 84.04/65.81       & 88.78/68.29       & 90.76/77.35       \\
SNARF \cite{chen2021snarf}  & 95.75/84.32       & 95.42/86.32       & 95.43/86.07       & {\bf 96.08}/85.47 & 95.57/85.01       & 96.05/82.50       & {\bf 95.69/82.11} & 94.44/83.41       & 95.35/83.41       & 95.22/84.91       \\
COAP                        & {\bf 95.97/85.35} & {\bf 95.84/87.62} & {\bf 95.57/86.82} & 95.98/{\bf 85.65} & {\bf 95.84/86.28} & {\bf 96.61/82.96} & 95.27/81.90       & {\bf 94.91/84.90} & {\bf 96.07/85.89} & {\bf 95.78/86.90} \\%[2pt]
% \midrule
% Neural-GIF~\cite{NGIF}      & \cmark & {\bf \phantom{0}22}  & 67.88/39.09       & 53.41/45.86       & 57.62/48.34       & 78.89/53.132      & 68.76/51.45       & 50.86/16.13       & 64.85/26.32       & 65.36/50.45       & 76.25/50.26       & 64.66/51.17       \\
% LEAP~\cite{LEAP:CVPR:21}    & \cmark & \phantom{0}35        & 88.53/67.05       & 90.43/77.85       & 89.84/76.15       & 88.18/64.79       & 91.33/77.09       & 74.66/35.30       & 83.65/53.82       & 84.04/65.81       & 88.78/68.29       & 90.75/77.35       \\
% COAP                        & \cmark & \phantom{0}75        & {\bf 95.83/84.09} & {\bf 96.95/90.57} & {\bf 96.93/90.36} & {\bf 96.59/87.16}  & {\bf 97.24/90.36}& {\bf 86.75/58.75} & {\bf 93.89/76.72} & {\bf 96.16/88.15} & {\bf 96.79/88.22} & {\bf 96.89/89.97} \\
\bottomrule
\end{tabular}
\caption{
    \textbf{Single-subject neural implicit models.} Comparison with NASA, LEAP and SNARF~\cite{chen2021snarf} on per-subject training.
    There results supplement results displayed in \tablename~\ref{tab:sota_single_person}.
    }
\label{tab:app:sota_single_person}
\end{table*}

\paragraph{Additional Results}
We provide additional qualitative results for the generalization experiment (Sec.~\ref{sec:exp}) in \figurename~\ref{fig:sup:sota_unseen_ppl} and additional quantitative results of two more baselines (NASA~\cite{nasa}, LEAP~\cite{LEAP:CVPR:21}) in \tablename~\ref{tab:app:sota_single_person} for the single-subject experiment.

\paragraph{Resolving Self-intersections (Sec.~\ref{subsec:selfint})}
For the baseline~\cite{pavlakos2019expressive, Tzionas:IJCV:2016}, we used default configuration parameters provided by the authors except for the collision weight, which we increased from $0.0001$ to $0.005$ for better performance. 
Our self-intersection procedure uses standard gradient-based optimization with a learning rate of $0.007$ and a total of $1300$ query points sampled (arbitrarily chosen) in the intersected volume of colliding bounding boxes.

\begin{figure}
    \begin{center}
        \includegraphics[width=1.0\linewidth,trim={200 150 400 550},clip]{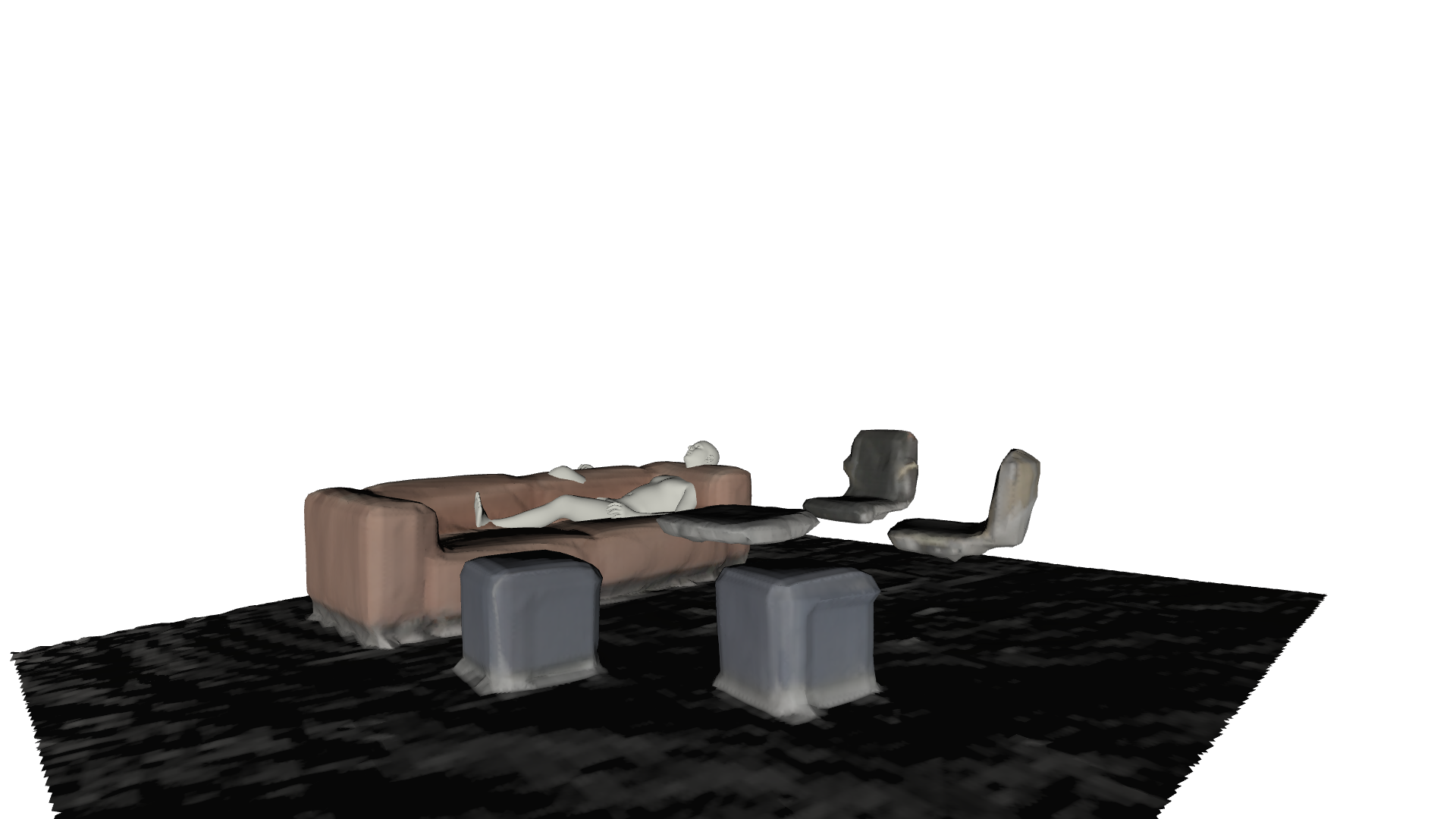}
    \end{center}
    \caption{\textbf{Limitation - resolving collisions with 3D environments.} 
    The optimization algorithm has difficulties resolving deep collisions with the environment as demonstrated here for an example from the PROX dataset~\cite{PROX:2019}.
    }
    \label{fig:app:coap_limit_opt}
\end{figure}

\paragraph{Resolving Collisions with 3D Environments (Sec.~\ref{subsec:scenepen})}
For the human-scene reconstruction pipeline, we use the optimization schedule from the PROX pipeline~\cite{PROX:2019} with the original weighting terms. 
Our proposed collision term is added to the final optimization loss and weighted by $100$. 
Please see the supplementary video for qualitative results. 
The optimization algorithm is sensitive to estimated joint locations and cannot resolve deep collisions with the environment (\figurename~\ref{fig:app:coap_limit_opt}).

\begin{table}
\centering
\scriptsize
\setlength{\tabcolsep}{3.0pt}
    \begin{tabular}{@{}lcccccccccc@{}}
        \toprule
        Steps: & 1\% & 5\% & 10\% & 15\% & 25\% & 30\% & 40\% & 50\% & 60\% & 70\% \\
        IOU:   & 96.86 & 96.93 & 96.96 & 96.96 & 96.97 & 96.98 & 96.98 & 96.98 & 96.97 & 96.96 \\
        \bottomrule
    \end{tabular}
    \caption{\textbf{Ablation of the bounding box size.}}
\label{tab:app:abl_box}
\end{table}
% \begin{table}[h!]
% \vspace{-6pt}
% \centering
% \scriptsize
% \setlength{\tabcolsep}{3.0pt}
% \renewcommand{\arraystretch}{1.0}
% \renewcommand{\bottomfraction}{0.0}
% \begin{tabular}{@{}lcccccccccc@{}}
%     % \toprule
%     Steps: & 1\% & 5\% & 10\% & 15\% & 25\% & 30\% & 40\% & 50\% & 60\% & 70\% \\
%     IOU:   & 96.86 & 96.93 & 96.96 & 96.96 & 96.97 & 96.98 & 96.98 & 96.98 & 96.97 & 96.96 \\
%     % \bottomrule
% \end{tabular}
% % \caption{\textbf{Single-subject neural implicit models.} }
% \label{tab:reb:bbsize}
% \vspace{-12pt}
% \end{table}
\paragraph{Ablation of the bounding box size}
We further study the impact of the size of the bounding boxes $B_k$ on model performance. 
We compute the uniform IoU in \tablename~\ref{tab:app:abl_box} for a varying number of up-sampling steps for the generalization experiment on the PosePrior dataset (Tab.~\ref{tab:sota_unseen_ppl} in the paper). 
Very tight boxes (less than 10\% of the original size) slightly degrade the representation quality, while the performance saturates at $15\%$. 
We decided to use a tight box in this range for the experiments simply because these bounding boxes are used to detect an initial set of potentially collided body parts for resolving self-intersections. If the boxes are too large, the initial set of candidates would be larger and slow down the optimization. 

\clearpage

% --- uncomment this to read the instructions
% \input{sec/X_instructions}

\end{document}